\documentclass{article}

\usepackage{microtype}
\usepackage{graphicx}
\usepackage{subfigure}
\usepackage{booktabs} 
\usepackage[dvipsnames]{xcolor,colortbl}
\usepackage{hyperref}

\usepackage[accepted]{icml2024}

\usepackage[compact]{titlesec}
\titlespacing{\section}{.5pt}{.5ex}{.5ex}
\titlespacing{\subsection}{.5pt}{.5ex}{.5ex}
\titlespacing{\subsubsection}{.5pt}{.5ex}{.5ex}

\usepackage[bf,small]{caption}

\usepackage{amsmath}
\usepackage{amssymb}
\usepackage{mathtools}
\usepackage{amsthm}

\usepackage[capitalize,noabbrev]{cleveref}

\theoremstyle{plain}

\theoremstyle{definition}

\theoremstyle{remark}

\usepackage{enumitem}
\setitemize{noitemsep,topsep=0pt,parsep=2pt,partopsep=0pt}

\icmltitlerunning{Neural Implicit Topology Optimization}

\begin{document}

\twocolumn[




\icmltitle{NITO: Neural Implicit Fields for Resolution-free Topology Optimization}



\icmlsetsymbol{equal}{*}

\begin{icmlauthorlist}
\icmlauthor{Amin Heyrani Nobari}{mit}
\icmlauthor{Giorgio Giannone}{mit}
\icmlauthor{Lyle Regenwetter}{mit}
\icmlauthor{Faez Ahmed}{mit}
\end{icmlauthorlist}

\icmlaffiliation{mit}{Department of Mechanical Engineering, Massachusetts Institute of Technology, Cambridge, MA, USA}

\icmlcorrespondingauthor{Amin Heyrani Nobari}{ahnobari@mit.edu}

\icmlkeywords{Machine Learning, ICML}

\vskip 0.3in
]

\printAffiliationsAndNotice

\begin{abstract}
Topology optimization is a critical task in engineering design, where the goal is to optimally distribute material in a given space for maximum performance. We introduce Neural Implicit Topology Optimization (NITO), a novel approach to accelerate topology optimization problems using deep learning.
NITO stands out as one of the first frameworks to offer a resolution-free and domain-agnostic solution in deep learning-based topology optimization. NITO synthesizes structures with up to seven times better structural efficiency compared to SOTA diffusion models and does so in a tenth of the time.
In the NITO framework, we introduce a novel method, the Boundary Point Order-Invariant MLP (BPOM), to represent boundary conditions in a sparse and domain-agnostic manner, moving away from expensive simulation-based approaches. 
Crucially, NITO circumvents the domain and resolution limitations that restrict Convolutional Neural Network (CNN) models to a structured domain of fixed size -- limitations that hinder the widespread adoption of CNNs in engineering applications.
This generalizability allows a single NITO model to train and generate solutions in countless domains, eliminating the need for numerous domain-specific CNNs and their extensive datasets. Despite its generalizability, NITO outperforms SOTA models even in specialized tasks, is an order of magnitude smaller, and is practically trainable at high resolutions that would be restrictive for CNNs. 
This combination of versatility, efficiency, and performance underlines NITO's potential to transform the landscape of engineering design optimization problems through implicit fields.

\end{abstract}

\section{Introduction}
\begin{figure}[ht!]    
    \centering
    \includegraphics[width=.9\columnwidth]{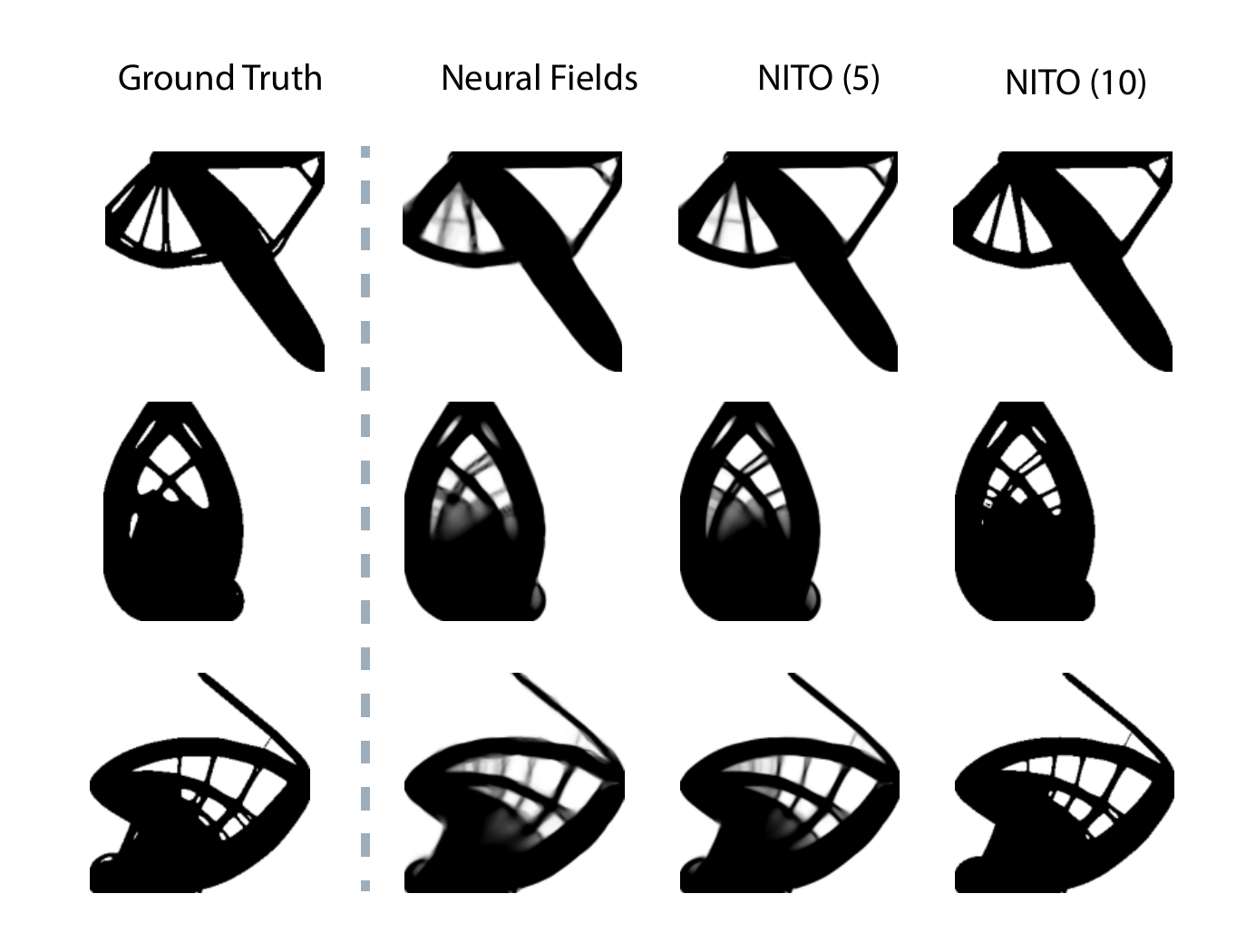}
    \caption{NITO framework leveraging BPOM.
    Left: ground truth obtained using a SIMP optimizer. Second From Left: the raw output of neural fields using BPOM. Right Two: the NITO framework output leveraging a few steps (5 and 10 steps) of optimization. We see that NITO is amenable to deep optimization and can generate high-quality, constraint-satisfying, and high-performance topologies with fast inference. See Appendix~\ref{appx:visualizations} for more visualizations.}
    \label{fig:deepopt}
\end{figure}
Recent years have witnessed remarkable advancements in vision and language domains, marked by a series of influential studies~\citep{child2020very, brock2018large, ho2020denoising, rombach2021high, devlin2018bert, brown2020language, raffel2020exploring}. These developments have fundamentally transformed our approach to processing multimodal and unstructured data. This transformation has been pivotal in fostering innovative generative methodologies for such data~\citep{ramesh2022hierarchical, nichol2021glide, blattmann2022retrieval, ramesh2021zero}. With the advent of these deep generative models~(DGM) and their success in language and vision, significant research has been developed specifically in the application of these deep generative models in scientific and engineering problems~\citep{regenwetter2022deep, song2023multi,HU2024103639,wu2021deepcad}.

Engineering design seeks to create designs that satisfy a set of constraints while maximizing functional performance. It is therefore easy to see how this kind of problem lends itself to optimization and is effectively addressed by optimization algorithms~\cite{boyd2004convex}, making them cornerstones in computational design. 
A classic objective for structural topology optimization~(TO) is to distribute a limited amount of material in a physical domain such that the resulting structure is as stiff as possible. This objective is called compliance minimization, but other objectives such as heat transfer are often considered. 
A popular TO method is the Solid Isotropic Material with Penalization method~\cite{SIMP1988, SIMP1992}~(SIMP). While methods such as SIMP are powerful engineering tools, they struggle with computational cost when solving large-scale problems due to their iterative nature and expensive per-step computations~\cite{sigmund_review}.

In recent years, many works have investigated the prospect of tackling topology optimization problems using machine learning~\citep{regenwetter2022deep}. Most of these approaches are built around DGMs trained on optimized topologies and conditioned on a set of loads and boundary conditions.~\citep{nie2021topologygan,maze2022topodiff,giannone2023aligning,giannone2023diffusing,HU2024103639}. The main justification for these approaches is significant time savings compared to TO solvers like SIMP and the ability to generate multiple near-optimal solutions. Indeed, any DGM able to quickly generate near-optimal solutions delivers immense value in the form of time and cost savings~\citep{Woldseth_2022}. 

However, compared to classic iterative optimizers like SIMP, pure data-driven approaches typically do not generate optimal topologies, despite being faster. This is primarily attributable to the fact that \emph{data-driven models typically focus on density estimation and remain agnostic to the physics of a given problem}. 
To address this, researchers have proposed to incorporate physics into these models, such as through guidance in diffusion models~\citep{maze2022topodiff} or aligning the trajectory of diffusion models with the optimizer trajectory~\cite{giannone2023aligning}.
Despite these efforts, the aforementioned approaches suffer from some important limitations. 

When applying TO most problems are defined in uniquely shaped domains and require different levels of detail~(i.e., resolution) depending on the application and the underlying problem. Most existing deep learning methods discretize the physical domain into pixels or voxels to leverage convolutions (CNNs) as basic building block~\citep{maze2022topodiff,giannone2023diffusing,giannone2023aligning,nie2021topologygan,HU2024103639}. 
This limits their generation capabilities to a specific resolution and shape (such as a square domain) -- a new resolution or aspect ratio necessitates an entirely new dataset and model. 
Moreover, these DGMs typically rely on physical fields (or approximations) as a representation of the boundary condition of the underlying problems~\citep{maze2022topodiff,giannone2023diffusing,giannone2023aligning,nie2021topologygan,constraintrepasme,HU2024103639}. These physical fields are represented as images to be handled by CNNs. Not only is this approach computationally expensive, particularly for high dimensionality, but it further exacerbates the lack of generalizability across problems and domains. This lack of generalizability is a common criticism directed at such methods by computational structure engineering~\citep{Woldseth_2022}. 

In this work, we propose a ``deep optimization'' approach that couples the strengths of deep learning models and optimization in an effective learning scheme. To avoid the limitations of CNN-based generative schemes, we propose an approach based on neural implicit fields, that can be applied to physical domains of any size, shape, and resolution. 
We also propose a generalizable scheme for representing sparse boundary conditions in numerical physics domains, eliminating the need for computationally expensive and domain-limiting physical field calculations, previously thought to be crucial for constraint representation~\citep{constraintrepasme}. 
Instead, we show that our computationally- and memory-efficient scheme is capable of representing sparse boundary conditions effectively, without loss of performance. Indeed our approach outperforms the state-of-the-art by more than 80\% while running 2.5-50 times faster than the fastest methods in the state-of-the-art.

We also re-implement an updated SIMP optimizer. This optimizer improves upon existing open-source Python implementations~\cite{hunter2017topy} using the latest advances in structural engineering optimizers for TO~\cite{Wang2021}. Leveraging this updated optimizer, we introduce a new dataset using our improved optimizer and retrain and benchmark previous state-of-the-art methods using this updated dataset.

\paragraph{Contributions.}
Our contributions can be summarized as follows:
\begin{itemize}


\item[\textbf{1.}] We introduce a novel framework for TO called ``Neural Implicit Topology Optimizer~(NITO),'' a model capable of generating near-optimal topologies.
NITO leverages ``Boundary Point Order-Invariant MLP~(BPOM),'' a simple, generalizable architecture to represent sparse physical boundary conditions, that can perform on par with computationally expensive physical fields. 

\item[\textbf{2.}] We show that NITO and BPOM enable resolution-free and generalizable generation of topologies and demonstrate that NITO can be trained on multiple resolutions/domains and generate topologies for multiple resolutions, making it one of the first resolution-free and truly generalizable frameworks for TO with deep learning addressing a major limitation in prior works. 

\item[\textbf{3.}] We empirically show that NITO is an effective framework, producing topologies with up to 80\% lower compliance errors than SOTA learning-based models, while requiring an order of magnitude fewer parameters which enables it to run 2.5-50x faster while retaining high performance. More notably we show that NITO is a generalizable framework, retaining similar performance metrics for higher resolution with the same architecture and number of parameters.

\item[\textbf{4.}] Finally, we implement a new SIMP optimizer for minimum compliance TO, which is based on the latest literature on the topic. We demonstrate that our new optimizer is more capable than the previous optimizer used as the gold standard in recent literature. The new optimizer produces topologies that are on average 3\% lower in compliance and runs up to 6x faster than prior Python implementations.
\end{itemize}

\section{Background}
This section will provide some background information on topology optimization and neural implicit fields. We will also review existing methods for optimal topology generation using deep generative models. For more related work and limitations of previous approaches, see Appendix~\ref{appx:related-work}.

\begin{figure}[ht]
    \centering
\includegraphics[width=\linewidth]{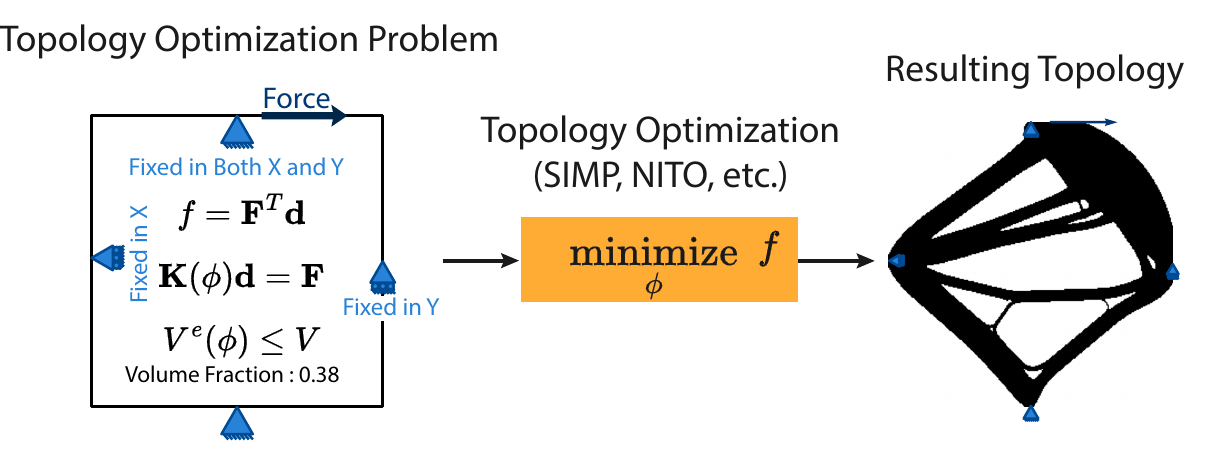}
    \caption{Topology Optimization. Given a domain, boundary conditions, loads, and volume fraction, TO aims to find the design variables $\phi$ that maximize performance (in this case minimizing compliance $f$) for the system, fulfilling all the prescribed constraints and respecting the underlying physics (Static Equilibrium).}
    \label{fig:to}
\end{figure}

\paragraph{Structural Topology Optimization}
Topology optimization~(TO) is a powerful computational method that identifies an efficient material layout, given a material budget, constraints, boundary conditions, and loads. In structural applications, TO often aims to minimize compliance (structural deformation) given a set of forces and supports (Fig.\ref{fig:to})~\cite{liu2014efficient}. A prominent technique in TO is the Solid Isotropic Material with Penalization~(SIMP) method, introduced by \citet{bendsoe1989optimal}. This method models material properties through a density field, where the density value signifies the amount of material in a specific region. The optimization iteratively simulates the system to evaluate the objective, then adjusts the density field based on the gradient of the objective score for the field. Mathematically, we represent the density distribution as $\rho (\phi)$, where $\phi$ is a set of design variables, each representing a value within the problem domain $\Omega$. The optimization task is structured as follows:
\begin{equation}
\begin{array}{cl}
\underset{\phi}{\operatorname{min}}  & f =\mathbf{F}^T \mathbf{d} \quad \\
\text { s. t. } & \mathbf{K}(\phi) \mathbf{d}=\mathbf{F} \\
& \sum_{e \in \Omega} \rho^e(\phi) v^e \leq V \quad\\
& \phi_{\min } \leq \phi_i \leq \phi_{\max } \quad \forall i \in \Omega
\end{array}
\end{equation}
The objective is to minimize the compliance $\mathbf{F}^T \mathbf{d}$, with $\mathbf{F}$ being the load tensor and $\mathbf{d}$ the nodal displacement, a solution to the equilibrium equation $\mathbf{K}(\phi) \mathbf{d}=\mathbf{F}$, where $\mathbf{K}(\phi)$ is the stiffness matrix contingent on the design variables $\phi$. The constraint $\sum_{e \in \Omega} \rho^e(\phi) v^e \leq V$ ensures the total volume does not exceed a maximum limit $V$~(often expressed as a fraction of the domain volume). The optimization seeks design variables within specified bounds ($\phi_{\min}$ and $\phi_{\max}$) for every element $i$ in the domain $\Omega$. By allowing continuous variation of design variables between 0 and 1, this formulation supports gradient-based optimization. Nonetheless, the expensive FEA simulation~(solving $\mathbf{K}(\phi) \mathbf{d}=\mathbf{F}$) at each iteration causes the overall optimization to be computationally intensive. Notably, solving the linear system of the equations scales cubically ~($O(n^3)$) with the size of the problem, in this case, the number of equations that must be solved~(two equations per node in 2D).

\paragraph{Implicit Neural Fields}
A neural field is a field that is partially or fully defined by a neural network~\citep{xie2022neural}. This neural network takes some form of coordinate representation in space $\mathbf{x} \in \mathbb{R}^n$ as input and outputs a set of values~(i.e. a measure of the desired field) $\Phi(\mathbf{x}) \in \mathbb{R}^m$:
\begin{equation}
    \Tilde{\Phi}(\mathbf{x}) = f_{\Theta}(\mathbf{x})
\end{equation}
where $f_{\Theta}$ is the neural network function given parameters $\Theta$. These neural fields have proven effective in representing audio~\citep{audionif}, images~\citep{imagenif,siren}, videos~\citep{videonif}, 3d objects~\citep{3dnif}, 3D scenes~\citep{nerf}, and many more applications~(see the review by \citet{xie2022neural} for a more comprehensive picture). It has also been shown that neural implicit representations can be directly optimized using gradient-based optimizers similar to SIMP to represent the optimal topology for any given problem~\citep{imto1,imto2}, which shows clearly that these implicit neural representations are indeed capable of generating density fields that represent \textit{individual} optimal topologies. In this work, we propose a conditional implicit neural field approach that is meant to generate \textit{different} optimal topologies based on conditions~(i.e., material budget, loads, supports).

\section{Methodology}
\begin{figure*}[!ht]
    \centering
\includegraphics[width=.9\linewidth]{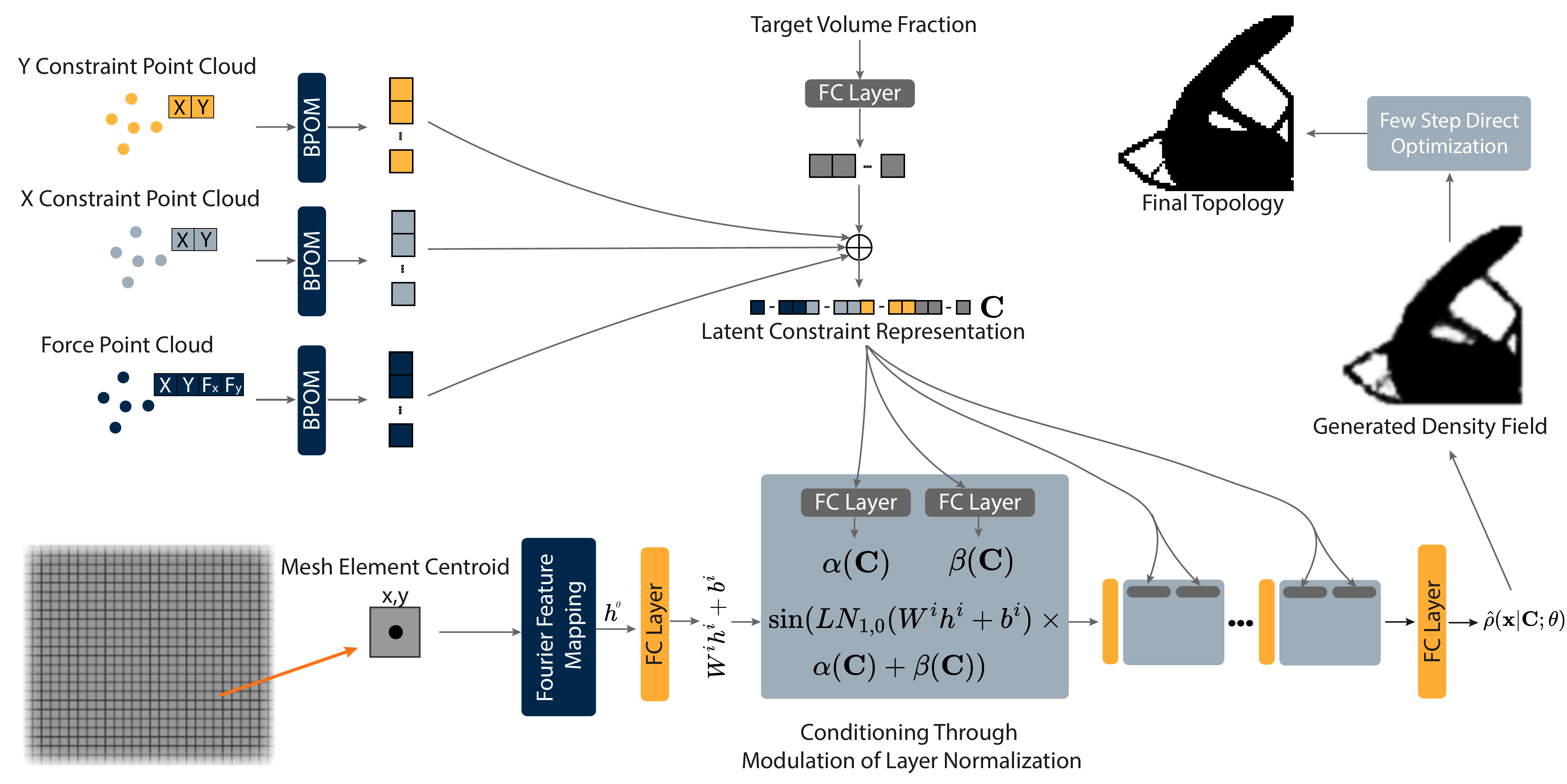}
    \caption{The NITO framework for topology optimization. The boundary conditions are processed as point clouds using the BPOM approach and a neural field is conditioned on these representations by modulating layer normalization based on the latent representation of the constraints to predict the optimal density field given the boundary conditions of the problem. Finally, the predicted density field is further refined through a few steps of direct optimization.}
    \label{fig:overview}
\end{figure*}
Our framework aims to enhance inference speed by eliminating iterative sampling for topology generation, ensuring linear scalability with sampled points. Additionally, it introduces a method for managing sparse conditioning in dense models without relying on fields. The framework, based on neural fields, overcomes size and shape constraints in images, enabling application to any spatial domain and improving generalizability in topology optimization (TO).



\subsection{NITO: Neural Implicit Topology Optimization}
Our main objective in this problem is to learn how to distribute material inside a domain such that the mechanical compliance of the resulting structure is minimized. As we discussed, the distribution of material is often described by a density field, which we denote as $\rho(\mathbf{x})$, where $\mathbf{x}$ is the coordinate of a given point in space and $\rho(\mathbf{x})$ is the density of material in that part of space. This formulation is highly compatible with neural fields. However, we do not intend to just learn a single distribution of material. Rather, we need to learn a conditional distribution based on the boundary conditions and volume ratio for a given problem. This means that we have to learn a conditional density field for a given problem:
\begin{equation}
    \hat{\rho}(\mathbf{x} | \mathbf{C};\theta) = f_\theta(\mathbf{x}, \mathbf{C}),
\end{equation}
where $f$ denotes the neural field function which is defined by the neural network architecture and $\mathbf{C}$ is a condition vector that includes information about the domain shape, boundary conditions, and desired volume ratio. In practice, this density field is desired to have a density of exactly 0 or 1 at any point, representing where material should or should not be placed. As such, the density fields that are generated by the SIMP optimizer for a given problem have binary values. 
In practice, it is more practical to formulate the problem as the probability of material being placed at any given coordinate. As such, the formulation of the objective can be written as:
\begin{equation}
\begin{split}
\mathcal{L}(\theta) = &-\mathbb{E}_{\mathbf{x}, \mathbf{C}} \left[ \rho(\mathbf{x} | \mathbf{C}) \log f_{\theta}(\mathbf{x}, \mathbf{C}) \right. \\
&\left. + (1 - \rho(\mathbf{x} | \mathbf{C})) \log (1 - f_{\theta}(\mathbf{x}, \mathbf{C})) \right]
\end{split}
\end{equation}
where $\rho(\mathbf{x} | \mathbf{C})$ is the probability of material existing at $\mathbf{x}$, which in this case is the same as the output of the SIMP optimizer. A detailed schematic of how the NITO framework operates is shown in Figure \ref{fig:overview}. We discuss implementation details of the neural fields and conditioning mechanism in Appendix~\ref{appx:implementation-details}.

\subsection{BPOM: Latent Constraint Representation}

Having discussed the neural field implementation, we now consider boundary conditions in discretized domains, as seen in TO problems. In particular, we discuss how arbitrary discrete boundary conditions can condition generative models in a sparse and generalizable way. This is particularly important for large domains, where defining constraints over grids is computationally cumbersome and memory-expensive.

\paragraph{Conditioning on Physics Fields}
Previous works~\citep{maze2022topodiff,nie2021topologygan,constraintrepasme} have argued that representing the boundary conditions of TO problems through simulation-derived physical fields such as stress and strain energy is \emph{the most effective} (or only) way to generate high-performing topologies using conditioning generative models. This has consequences for the generalizability of these methods since the fields often have to be handled by CNN models which are fixed to one resolution and domain.
\begin{figure}[!ht]
    \centering
\includegraphics[width=\linewidth]{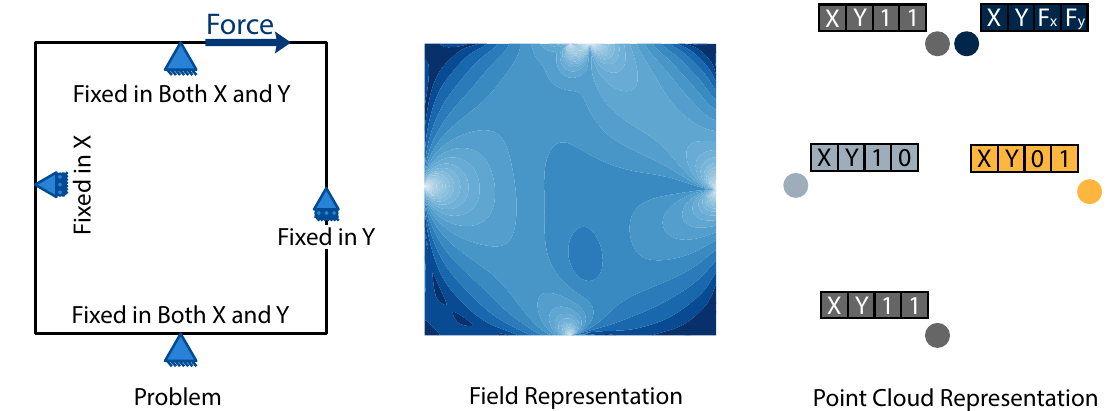}
    \caption{Comparison of field-based such as stress fields (middle), and point-cloud-based (right) representations, given a TO problem (left). Unlike the expensive and domain-limiting iterative FEA method, the point clouds offer a generalizable and memory efficient, representation of the boundary conditions.}
    \label{fig:bc}
\end{figure}
We believe that field-based conditioning's popularity is largely attributable to a conditioning mechanism limitation. In most existing studies, authors have only considered conditioning models at the input of the first layers. Using this approach, field-based conditioning was the only effective method. However, it is well understood that this input-only conditioning allows models to ignore the input conditions to a great extent~\citep{perez2017film}, and that more efficacious approaches exist. Therefore, in this paper, we propose a simple and effective model to represent conditions and incorporate them more effectively while enabling more generalizable and widespread use.



\paragraph{Constraints as Point Clouds}
In our approach, we aim to represent the boundary conditions of TO problems in a manner that is \emph{agnostic to the problem domain}. This enables our method to easily generalize to different domains without needing to train separate models on individual datasets for each domain shape and resolution. 
In all problems, conditioning is based on loads, displacement constraints, and volume fraction (see Figures \ref{fig:bc} and \ref{fig:to}). Since a problem may have arbitrarily many loads or boundary conditions, we need a mechanism that can take all loads and boundary conditions for a problem and reduce them to a single latent representation. To do this, we treat these sparse conditions as a set of three point clouds -- one for loads, x supports, and y supports (in higher dimensional problems, more point clouds can be used)~(see Figures \ref{fig:overview} and \ref{fig:bc}). Then we process each point cloud using ResP Layers proposed by~\citet{ma2022rethinking} in their point cloud model, PointMLP. 
However, unlike PointMLP, we did not find a need for the geometric affine module (as proposed by the authors for complex point cloud geometries), given the simpler geometric nature of our sparse boundary condition point clouds. 

\paragraph{Order-Invariant Aggregation}
As discussed, the point clouds can have any size. We therefore take the output of the point cloud model and perform order-invariant pooling to reduce every point cloud into one single vector representing the boundary conditions. We do this by computing minimum, maximum, and average pooling and concatenating the three pooling results. In the end, the three point clouds are each reduced, then concatenated into one vector. This vector is then input to the modulated layer normalization mechanism in our conditional neural fields~(see Appendix \ref{appx:implementation-details} and Figure \ref{fig:overview}). We call this boundary condition representation Boundary Point Order-Invariant MLP~(BPOM) and we show that this representation works just as well as physical fields. Finally, for the volume fraction condition (which is always a single floating point value), we use a single fully-connected layer and concatenate its output to the BPOM outputs of the boundary condition point clouds~(Figure \ref{fig:overview}).


\subsection{Few-Step Direct Optimization On Neural Field}
Despite the success of generative models and other deep learning techniques for TO, we observe that the performance of these models always lags behind the optimization baseline. To address this, \citet{giannone2023aligning} proposed applying a few steps~(5-10 steps) of optimization to the outputs of deep generative models with SIMP~(compared to 200-500 iterations for full optimization). This allows SIMP to rapidly hone in on more accurate structures for the given boundary conditions using near-optimal generated topologies as a starting point. We see this `optimization-seeding' approach as a crucial step in topology generation using generative models and inherit it in our approach. To ensure that the optimization itself is performed optimally, we implement the SIMP method based on the latest research on TO~\citep{WangZhaoZhou2021} and use this solver for our experiments and training. This completes the NITO framework as a ``deep optimization'' approach~(Figure \ref{fig:overview}).

\section{Experiments}
In this section, we present an assortment of experiments to compare the performance of NITO to existing state-of-the-art models in the literature.
With these experiments, we provide compelling evidence that:
\begin{itemize}
\item[\textbf{1.}] NITO is a scalable, resolution-free paradigm that outperforms convolution-based methodologies, with a significantly smaller parameter count.

\item[\textbf{2.}] NITO is fast and can generate constraint-satisfying, high-performance topologies much faster than a state-of-the-art SIMP optimizer, and

\item[\textbf{3.}] NITO merges the power of deep learning models as an efficient first stage, and the precision and guarantees of optimization as a reliable second stage to form an effective `deep optimization' framework. Deep optimization is a promising research direction for large-scale adoption and deployment of deep learning methodologies in engineering design.
\end{itemize}

For more experiments and discussion, see Appendix~\ref{appx:additional-experiments}.
See Appendix~\ref{appx:visualizations} for extensive visualizations.

\subsection{Experimental Details}
Before presenting results, we will discuss the different aspects of our experiments in this section.

\paragraph{Data}
We created a dataset of SIMP-optimized topologies for the 64x64 and 256x256 unit square domains. We include loading conditions, boundary conditions, volume fractions, and stress and strain energy fields in our dataset. Noting that previous works used older implementations of SIMP, specifically ToPy~\citep{hunter2017topy}, which lacks recent improvements and uses a slower solver, we developed a new SIMP optimizer in Python. This implementation utilizes the most current and efficient SIMP algorithm available~\citep{WangZhaoZhou2021}. We then re-created the 64x64 dataset from \citet{maze2022topodiff} with our solver, finding significant improvements in optimality. For fairness in comparison, we retrained the TopoDiff model\citep{maze2022topodiff}~(the best-performing model in literature) on our new dataset, while also acknowledging the original performance metrics reported by other authors. Further details on the dataset and solver are included in Appendix \ref{appx:data}.

\paragraph{Evaluation Metrics}
We evaluate the models in terms of performance (i.e., minimum compliance) and constraint satisfaction.
First, to measure how well the models perform in compliance minimization, we measure compliance error~(CE), which is calculated by subtracting the compliance of a generated sample from the compliance of the SIMP-optimized solution for the corresponding problem. We also measure the volume fraction error~(VFE), which quantifies the absolute error between the generated topology's actual volume fraction and the target volume fraction for the given problem. Beyond these base performance metrics, we quantify inference time. The speed of the SIMP optimizer is also benchmarked for comparison.
We remove outlying samples, which have extremely high compliance errors~(above 1000\%) due to models failing to place material in locations where the load is applied. We do this for all models including other SOTA models that we compare against, as is common practice~\citep{maze2022topodiff,giannone2023aligning,nie2021topologygan}.

\paragraph{Setup}
We train NITO for 50 epochs in three different scenarios: (1) We train on 30,000 topologies with a resolution of 64x64. (2) We train on 60,000 256x256 topologies. (3) We train on both datasets simultaneously to demonstrate the generalizability of our approach. To compare our approach to the state of the art, we also train TopoDiff on the 64x64 in the manner the original authors did~\citep{maze2022topodiff}. Notably, TopoDiff is currently the state-of-the-art model and outperforms all existing models in literature~\citep{giannone2023aligning}. Training a diffusion model like TopoDiff on large images of size 256x256 is rather difficult and can be prohibitive for most practitioners given the memory requirements and time needed to train these models. Despite this, we train TopoDiff on the 256x256 dataset for 500,000 steps and report the results. 

\subsection{Performance}
\begin{table}[ht!]
    \centering
    \caption{Quantitative Evaluation on 64x64 datasets. All TopoDiff variants have been re-trained on optimized topologies obtained using our improved SIMP optimizer. $^*$TopologyGAN and cDDPM results from~\citet{maze2022topodiff} and~\citet{giannone2023aligning}. w/ G: using a classifier and regression guidance. FS-SIMP: Few-Steps of optimization. CE: Compliance Error. VFE: Volume Fraction Error. 
    NITO achieves SOTA performance on topology optimization in terms of compliance and volume fraction error.}
    \setlength{\tabcolsep}{3pt}
    \resizebox{\linewidth}{!}{
    \begin{tabular}{l c cccc}
\toprule
Model              & FS-SIMP &   CE \% Mean  &  CE \% Med  &  VFE \% Mean  &  VFE \% Med \\
\midrule
TopologyGAN$^*$   &  - & 48.51 & 2.06 & 11.87 & - \\
cDDPM$^*$     &  -   &    60.79 & 3.15 & 1.72  & - \\
TopoDiff        &  -   &      3.23   &    0.45  &    1.14&    0.94 \\
TopoDiff w/ G        &  -  &    2.59&   0.49  &     1.18    &    0.99 \\
\midrule
NeuralField    & -  &     6.28   &   \colorbox{SpringGreen}{0.38}  &    1.37     &   \colorbox{SpringGreen}{0.75}  \\
\textbf{NITO (ours)}& 5 &     \textbf{0.81}    &  \textbf{0.098}  &    \textbf{0.50}  &     \textbf{0.32}  \\
\textbf{NITO (ours)}  & 10  &  \textbf{0.48}   &  \textbf{0.067}   &  \textbf{0.41}     &   \textbf{0.28}  \\
\bottomrule
\end{tabular}
}
    \label{tab:performance}
\end{table}

\begin{table}[ht!]
    \centering
    \caption{Quantitative Evaluation on 256x256 datasets. The columns are the same as Table \ref{tab:performance}. NITO with BPOM and a few steps of optimization is effective at generating topologies with high performance irrespective of the considered problem resolution. On the other hand, the performance of TopoDiff has degraded so much that raw neural fields~(highlighted in green) outperforms it.}
    \setlength{\tabcolsep}{3pt}
    \resizebox{\linewidth}{!}{
    \begin{tabular}{l c cccc}
\toprule
Model              & FS-SIMP &   CE \% Mean  &  CE \% Med  &  VFE \% Mean  &  VFE \% Med \\
\midrule

TopoDiff          &  -   &  16.62    &    0.59  &    2.92&    2.29\\
\midrule
\rowcolor{SpringGreen} NeuralField    & -  &     7.52   &   0.72  &    1.39     &   0.89  \\
\textbf{NITO (ours)}  & 5  &   \textbf{0.38}    &  \textbf{0.035}  &    \textbf{0.56}  &     \textbf{0.37}  \\
\textbf{NITO (ours)}& 10 &      \textbf{0.10}   &  \textbf{0.011}   &  \textbf{0.38}     &   \textbf{0.28}  \\

\bottomrule
\end{tabular}
}
    \label{tab:256performance}
\end{table}

In Table \ref{tab:performance} we present a comprehensive quantitative evaluation of SOTA models for topology optimization on the 64x64 datasets. TopologyGAN and cDDPM show higher values in CE \% Mean and VFE \% Mean compared to other models, which is expected since TopologyGAN is a much older benchmark and cDDPM is a naive conditioning of diffusion models and is not expected to perform well in such a complex problem. A vanilla neural fields model without optimization does not perform as well as TopoDiff on average, however, the median performance of the neural fields model is better than all other baselines. This is due to a small number of problems that incur higher compliance errors, while the majority of samples outperform those generated by the state-of-the-art as reflected by the median values~(highlighted in green on table \ref{tab:performance}). A similar pattern is apparent in volume fraction error scores as well, however, the mean values are not as far apart in this realm. Unambiguously though, Table \ref{tab:performance} demonstrates that our NITO framework achieves a significant leap in performance compared to the SOTA. Neural fields start with average compliance errors of more than double TopoDiff. However, after even 5 steps of direct optimization, NITO outperforms TopoDiff and other methods by a large margin. When taking 10 steps of direct optimization, NITO achieves results that are very close to the optimizer, which takes up to 500 iterations to converge.

Next, we proceed to test NITO and TopoDiff on the much more challenging dataset of 256x256 topologies. Table~\ref{tab:256performance} presents the results of this study for NITO and TopoDiff~(the leading SOTA approach). In this case, we train TopoDiff with a larger model size to allow for effective learning on the 256x256 images. Despite this, we see a significant performance degradation of TopoDiff with increased resolution. This performance decrease is so severe that the vanilla neural fields~(highlighted in green in table \ref{tab:256performance}) outperform TopoDiff. This is while NITO achieves results on par with or slightly better than the 64x64 tests, all while using the same model size architecture and training.

\subsection{Direct Optimization Study}
\begin{table}[ht!]
    \centering
    \caption{In this table, we present the results of applying direct optimization on different approaches. This table demonstrates that the topologies predicted by neural fields are closer to the optimal. This indicates that despite lacking some finer details, the few-step optimizer quickly converges to a very good solution. This is contrasted with the topologies generated by TopoDiff which are detailed, but further from an optimal solution, reducing the benefits of additional optimizer steps.
    }
    \setlength{\tabcolsep}{3pt}
    \resizebox{\linewidth}{!}{
    \begin{tabular}{lcccc}
\toprule
Model & FS-SIMP & CE \% Mean  &  CE \% Med  &  VFE \% Mean  \\
\midrule
TopoDiff  &  -  & 3.23 (-)&  0.45 (-)&   1.14 (-)\\
TopoDiff    &  5  &    3.55 \textcolor{Red}{(+9.91\%)}&    0.42 \textcolor{ForestGreen}{(-6.67\%)}&    0.67 \textcolor{ForestGreen}{(-41.2\%)}\\
TopoDiff     &  10 &   1.38 \textcolor{ForestGreen}{(-57.3\%)}&    0.33 \textcolor{ForestGreen}{(-26.7\%)}&    0.45 \textcolor{ForestGreen}{(-60.5\%)}\\
\midrule
TopoDiff w/ G        &  -  &    2.59 (-)&   0.49 (-)&     1.18 (-)\\
TopoDiff w/ G   &  5  &   2.24 \textcolor{ForestGreen}{(-13.5\%)}&   0.44 \textcolor{ForestGreen}{(-10.2\%)}&   0.69 \textcolor{ForestGreen}{(-41.5\%)}\\
TopoDiff w/ G      &  10  &   1.05 \textcolor{ForestGreen}{(-59.5\%)}&    0.32 \textcolor{ForestGreen}{(-34.7\%)}&   0.45 \textcolor{ForestGreen}{(-61.9\%)}\\
\midrule
NeuralField    & -  &  6.28 (-)&   0.38 (-)&    1.37 (-)\\
\textbf{NITO (ours)}     & 5  &    \textbf{0.81} \textbf{\textcolor{ForestGreen}{(-87.1\%)}}&  \textbf{0.098} \textbf{\textcolor{ForestGreen}{(-74.2\%)}}&    \textbf{0.50} \textbf{\textcolor{ForestGreen}{(-63.5\%)}}\\
\textbf{NITO (ours)}& 10  &  \textbf{0.48} \textbf{\textcolor{ForestGreen}{(-92.4\%)}}&  \textbf{0.067}\textbf{\textcolor{ForestGreen}{(-82.4\%)}}&  \textbf{0.41} \textbf{\textcolor{ForestGreen}{(-70.1\%)}}\\
\bottomrule
    \end{tabular}
}
    \label{tab:fewstep}
\end{table}

In Table \ref{tab:fewstep}, we provide performance metrics for NITO and TopoDiff with direct optimization. Optimization of topologies generated by TopoDiff yields a much smaller performance boost compared to NITO and even reduces performance in some cases. As a result, compared to TopoDiff with 10 iterations of direct optimization, NITO outperforms with only 5 steps and widens the gap with 10. Importantly, NITO, despite being a much smaller model without insight from physical fields, generates topologies that are closer to optimal than CNN-based generative methods like TopoDiff. However, the topologies generated by neural fields lack more refined details~(see Figure \ref{fig:deepopt}), which is quickly resolved through a few iterations of optimization~(5-10 iterations compared to 500 iterations of the full optimization). This establishes NITO as a more robust, generalizable, and efficient framework for optimal topology generation.

\subsection{Inference Speed \& Efficiency}
\begin{table}[!ht]
    \centering
    \caption{
    Average Inference time for different problem resolutions. For the baselines, the inference time is estimated without considering the few steps of SIMP to improve performance. We include 10 SIMP iterations when computing NITO inference time. NITO is resolution-free, i.e. we can leverage the same small model for 64x64, 256x256, and any intermediate resolution. These times are measured using an RTX 4090 GPU and an Intel Core i-9-13900K CPU. We run the SIMP optimizer for 300 iterations.
    We compute the conditioning fields for TopoDiff using our fast solver, greatly speeding up inference for the baselines.
    }
    \setlength{\tabcolsep}{3pt}
    \resizebox{\linewidth}{!}{
    \begin{tabular}{lcc|cc}
 & \multicolumn{2}{c}{64x64 Resolution}& \multicolumn{2}{c}{256x256 Resolution}\\
    \toprule
    Model & Params (M)  & Inference (s)  & Params (M)  &Inference (s)  
\\
    \midrule
     TopoDiff    &   121     &  1.86  & 553     & 10.81  
\\
     TopoDiff w/ G &  239 & 4.79  & 1092 &   22.04
\\
     DOM           &   121   &  0.82   & 553   & 7.82   
\\
     NeuralFields &   \textbf{22}   &   \textbf{0.005}     & \textbf{22}   &\textbf{0.16}     
\\
     SIMP~\cite{hunter2017topy} &  - &  18.12 & - &316.02 
\\
     SIMP (our implementation) & -& 3.45  & -& 69.45 
\\
    \textbf{NITO (ours)}  & \textbf{22} & \textbf{0.14}  & \textbf{22} &\textbf{2.88}  \\
    \bottomrule
    \end{tabular}
    }
    \label{tab:inference}
\end{table}
In Table \ref{tab:inference} we present a comparative analysis of the generation speed of NITO against classic optimization methods and generative models, specifically focusing on inference time across different problem resolutions (64x64 and 256x256). The main objective of these kinds of deep learning schemes for data-driven design is to generate topologies quickly, in a manner that is faster than conventional optimization. In Table \ref{tab:inference} we see that in both 64x64 and 256x256 resolutions, NITO is significantly faster than the state-of-the-art models. Specifically for the 64x64 resolution, NITO is 83\% faster than the fastest state-of-the-art model DOM~\cite{giannone2023aligning}, and 97\% faster than our fast SIMP implementation. 
This demonstrates the speed and efficiency of NITO which is orders of magnitude faster than the SOTA, while simultaneously generating superior topologies. These gaps are widened in higher resolution tests with NITO  significantly outperforming baselines in inference speed, compliance error, and volume fraction error. NITO's superior scaling with resolution reflects the performance degradation of CNN-based models at high resolutions.

Table \ref{tab:inference} highlights another important metric, which is the number of parameters each model uses in their architecture. NITO can be trained on both image resolutions with the same number of parameters, 22 million, achieving SOTA performance, while CNN-based models have to be made larger and still face significant performance degradation as we demonstrated in prior sections. This is further evidence of the efficiency and scalability of NITO. 
We train NITO for both 64x64 and 256x256 resolution for the same number of steps while sampling the same number of points for each batch during training, which means that the model trains roughly for the same amount of time and with the same memory requirements. In fact a {\emph{single consumer GPU}~(we use an RTX 4090) is enough to train NITO. On the contrary, diffusion models like TopoDiff or DOM must grow to match larger resolutions and therefore require more memory and time to run and train. For resolutions above 256x256 or 3D TO problems, these frameworks can be impractical to train or run for most practitioners. In contrast, NITO is built to generalize to different domains/resolutions without issue, allowing for practical training of large problems with consumer-level computational resources. In the following, we specifically showcase the versatility of NITO when handling different domains.

\subsection{Resolution-Free Generalization}
\begin{table}[ht!]
    \centering
    \caption{Quantitative Evaluation on both datasets using models trained on mixed resolution data and different resolutions. NITO variants in this table use 10 steps of direct optimization. The results show that NITO trained on mixed-resolution data performs well across different resolutions.}
    \setlength{\tabcolsep}{3pt}
    \resizebox{\linewidth}{!}{
    \begin{tabular}{l lccc}
\toprule
Model               &Train Res.&   CE \% Mean  &  CE \% Med  &  VFE \% Mean \\
\midrule
 \multicolumn{5}{c}{64x64 Test}\\
 \midrule

Neural Fields&64& 6.28& 0.38& 1.37\\
NITO&64& 0.48& 0.067& 0.41\\[0.07in]
Neural Fields&256& 10.5& 0.70& 1.67\\
NITO&256& 0.51& 0.085& 0.50\\[0.07in]
 Neural Fields& 256,64& 7.11& 0.39& 1.36\\
 NITO& 256,64& 0.74& 0.069& 0.52\\
\midrule
\multicolumn{5}{c}{256x256 Test}\\
\midrule
 Neural Fields& 64& 10.47& 0.81& 1.46\\
NITO&64& 0.26& 0.10& 0.34\\[0.07in]
Neural Fields&256& 7.52& 0.72& 1.39\\
NITO&256& 0.10& 0.011& 0.38\\[0.07in]
 Neural Fields& 256,64& 8.47& 0.72& 1.43\\
 NITO& 256,64& 0.13& 0.012& 0.40\\
\bottomrule
\end{tabular}
}
    \label{tab:res-free}
\end{table}
A foundational pillar of NITO is its generalizability and adaptability to multiple resolutions and physical domains. We specifically conduct a study to showcase this generalizability and resolution-free adaptability. As mentioned before, we trained NITO on both 64x64 and 256x256 data separately, but here we also present results for training NITO on both the 64x64 and 256x256 datasets simultaneously. Then we use all three models to test on both the 64x64 test dataset and the 256x256 test dataset. These results are presented in detail in Table \ref{tab:res-free}. We see that the model trained on both datasets simultaneously performs on par and better than other SOTA models on both the 256x256 data and the 64x64 data. This proves NITO's capacity to be trained on multiple domains and perform just as well on both domains. 
Leveraging NITO's flexibility and BPOM's adaptability, models trained at one resolution can be effectively applied to tasks at different resolutions. This capability is demonstrated by testing NITO models trained on 64x64 data with 256x256 resolution tasks, and vice versa, showcasing the model's resolution-free generalization.
Remarkably, NITO's performance is consistent across resolutions: it performs similarly when trained on lower-resolution data compared to higher or mixed-resolution data, and likewise when tested on higher-resolution or lower-resolution problems. This underscores NITO's exceptional adaptability, indicating that its architecture not only supports training across multiple domains but also facilitates the transfer of learning from one domain to another, provided the problems share related distributions. This ability signifies a critical advantage of such frameworks---the potential to train on cost-effective low-resolution data and subsequently refine the models for higher resolutions with minimal additional data and reduced time.

\section{Conclusion \& Limitations}

In this paper, we introduce Neural Implicit Topology Optimization (NITO), a novel resolution-free and domain-agnostic deep-learning framework for topology optimization. Leveraging our proposed Boundary Point Order-Invariant MLP (BPOM) to represent boundary conditions, NITO eliminates the need for costly physical fields, overcoming the resolution and domain limitations of CNNs. NITO significantly outperforms state-of-the-art models in multiple resolutions and domains. Notably, it does so with remarkable speed and efficiency, requiring far fewer parameters. NITO's scalability and generalizability offer a robust foundation for future models in topology optimization and other physics-based problems, conferring a solution to high-dimensional problems that were previously insurmountable with CNN-based methods.

\paragraph{Limitations} 
While NITO presents significant advancements in topology optimization, it is not without potential shortcomings. 
Firstly, NITO is not a generative model, which means that the outputs of NITO are deterministic and not diverse for a given set of constraints. This limits NITO's ability to generate solutions that enable exploration of the design space. This characteristic could potentially hinder NITO's efficacy in addressing problems outside its training distribution, given the lack of a generative mechanism that enhances performance in entirely new tasks~(See Appendix \ref{appx:additional-experiments} for a broader discussion on this).  Future work could be devoted to addressing this matter, potentially by leveraging recent advances in generative frameworks for neural implicit fields~\citep{you2023generative,kosiorek2021nerfvae}. The second limitation arises from the fact that NITO requires a small number of direct optimization steps, and without this, the neural field-generated topologies with BPOM lack the necessary details. 
Therefore, future improvements should aim at refining NITO's training procedures and architecture. Such enhancements would reduce the reliance on direct optimization, empowering the neural fields to produce topologies with greater detail and precision.

\clearpage
\small
\section*{Impact Statement}
While the introduction of Neural Implicit Topology Optimization (NITO) represents a significant breakthrough in topology optimization, bringing efficiency and resolution-free solutions, it is crucial to consider the potential implications for society and the engineering field. 

Firstly, the automation and efficiency of NITO could lead to significant shifts within the engineering sector. The traditional roles of engineers, especially those involved in iterative design and optimization processes, may evolve or diminish as automation becomes more prevalent. This could lead to a need for re-skilling or up-skilling of the workforce to ensure that engineers remain relevant and can work synergistically with AI-driven systems like NITO.

Moreover, like with most advanced technologies, there is a risk of over-reliance on our framework. If engineers rely too heavily on NITO without fully understanding its internal workings, there might be a knowledge gap that could lead to oversights in design or an inability to troubleshoot and innovate beyond what the model offers. The ``black box'' nature of deep learning models can sometimes lead to a lack of interpretability, making it difficult to diagnose errors or understand the model's decision-making process in depth.
This is crucial as TO is often used in safety-critical components, from brackets to bridges. Deploying models without testing their performance using experiments creates a risk.


\bibliography{biblio}
\bibliographystyle{icml2024}

\newpage
\normalsize
\appendix
\onecolumn

\section{Related Work}
\label{appx:related-work}
In this section, we will discuss some of the related works in literature that work towards similar goals to our work.

The recent successes of Deep Learning (DL) in the field of vision have sparked significant interest in adapting these techniques to engineering applications. Specifically, DL methods have been applied to direct design \citep{Abueiddaetal2020, AtesGorguluarslan2021, behzadi2021real, MaZeng2020, UluZhangKara2016}, accelerating the optimization process \citep{Bangaetal2018, JooYuJang2021, Kallioras2020, SosnovikOseledets2017, Xueetal2021}, enhancing shape optimization post-processing \citep{Hertleinetal2021, Yildizetal2003}, achieving super-resolution \citep{Elingaard2021, Napieretal2020, Yooetal2021}, conducting sensitivity analysis \citep{AuligOlhofer2013, AuligOlhofer2014, Barmadaetal2021, Sasaki2019}, generating 3d topologies \citep{kench2021generating, sabiston20203d, behzadi2021real}, among others \citep{chen2022concurrent, chandrasekhar2022gm, ChandraSuresh2021, DengTo2021}. Generative Models, in particular, have been highlighted for their capability to increase design diversity in engineering design \citep{ahmed2016discovering, nobari2021creativegan, nobari2021pcdgan, nobari2022range}. Within Topology Optimization (TO), efforts have been concentrated on enhancing diversity through data-driven approaches \citep{JiangChenFan2021_nano2, RawatShen2018, RawatShen2019, KESHAVARZZADEH2021102947, SunMa2020}. Additionally, Generative Models are being utilized for TO problems, conditioning on constraints such as loads, boundary conditions, and volume fraction for structural cases, to directly generate topologies from a training dataset of optimized topologies, apply superresolution methods to increase fidelity \citep{yu2019deep, Li2019}, and use filtering and iterative design methods to improve quality and diversity \citep{oh2018design, oh2019, berthelot2017began, Fujita2021Design}. Proposals for methods addressing 3D topologies have also been put forward \citep{Behzadi_Ilies2021, KESHAVARZZADEH2021102947}. Recent advancements include GAN \citep{nie2021topologygan} and DDPM-based approaches \citep{maze2022topodiff}, which have shown promise in modeling the TO problem by conditioning on constraints and incorporating physical information. A detailed review and critique of this evolving field is provided by \citep{woldseth2022use}.

\paragraph{Deep Learning For Topology Optimization} The limitations associated with conventional optimization have motivated a significant body of work around deep learning approaches for topology optimization~(TO). Most relevant to our work are other methods that perform the task in an end-to-end manner, meaning they take constraints and boundary conditions as input and produce near-optimal topologies to minimize compliance~\citep{oh2019, Sharpe2019, 10.1007/s00158-020-02748-4, parrott2022multi}. For example, \citet{Yu2018} propose an auto-encoding GAN approach that uses an autoencoder for topology generation and a GAN for super-resolution which is applied simultaneously. \citet{Rawat2019}, \citet{Li2019} take a similar approach but use GANs for both initial generation and super-resolution. This is while \citet{Sharpe2019}, \citet{Nie2021}, and \citet{gantl} propose directly generating topologies using conditional GANs and \citet{Zhang2022} introduce U-Net-based architectures for TO.  More recently, researchers have looked at using implicit neural fields to generate topologies. Specifically, \citet{HU2024103639} use implicit neural representations to produce topologies in their approach called IF-TONIR. In their approach, however, the authors use stress and strain fields for their representations of boundary conditions which rely on CNNs and somewhat reduce the generalizability of the implicit neural fields' versatility.

\paragraph{Diffusion-based Topology Optimization} 
New methods that utilize diffusion models have recently achieved significant performance improvements over the aforementioned methods~\citep{maze2022topodiff,giannone2023diffusing,giannone2023aligning}. \citet{maze2022topodiff} were the first to do so and demonstrated that diffusion models are significantly better than GANs at optimal topology generation. They also showed that introducing guidance based on a regression model for predicting compliance and a classifier trained to identify floating material significantly improves the performance of such models~\citep{maze2022topodiff}. Like iterative optimization, however, diffusion models are slow, requiring the models to be run as many as 1000 times to generate a sample. This puts their value into question when the time savings are not notable compared to the optimizer itself. \citet{giannone2023aligning} propose aligning the diffusion models denoising with the optimizer's intermediate designs as a way to reduce the number of iterations needed for the diffusion model and show that they can indeed reduce the number of sampling steps significantly. 

\paragraph{Limitations}
Despite the upsides of the best-performing models which are diffusion-based, they lack speed and are all based around CNNs, which means that they treat the problem as images. Furthermore, the mechanism they use for conditioning based on boundary conditions is field-based, meaning these methods compute some physical or energy field to represent the boundary conditions of the problem, treat these fields as images, and employ CNN-based architectures to handle them. 
This method introduces two limitations. These models generate images instead of density fields, restricting them to a specific problem domain and resolution~(e.g. a square-shaped domain treated as 64x64 images) and making them unsuitable for different domain shapes and resolutions. Additionally, the use of image-based fields for boundary conditions further narrows their generalizability. This approach fails to capture boundary conditions in their raw form, limiting applicability to problems with boundary conditions defined for the exact domain they are trained on and excluding problems with irregular boundary conditions.


\section{Additional Experiments}
\label{appx:additional-experiments}
Here we will discuss some further experiments and discuss some limitations of our approach.

\subsection{Out of Distribution Experiments}
\begin{table}[ht!]
    \centering
    \caption{Quantiative Evaluation on out-of-distribution 64x64 datasets. w/ G: using a classifier and/or regression guidance. FS-SIMP: Few-Steps of optimization. CE: Compliance Error. VFE: Volume Fraction Error.}
    \setlength{\tabcolsep}{4pt}
    \begin{tabular}{ll ccccc}
\toprule
Model      & Train Res.        & FS-SIMP &      CE \% Mean  &  CE \% Median  &  VFE \% Mean  &  VFE \% Median \\
\midrule
\texttt{out-of-distro} \\
\midrule
NeuralField & 64   & -  &     73.24      &     12.81   &    8.12   &     6.87  \\
TopoDiff  & 64  &  -     &           8.57    &     1.14  &    1.14     &   0.97 \\
TopoDiff w/ G & 64 &  -             &     \textbf{7.79}   &    1.26   &    1.21    &    1.00 \\
\midrule
TopoDiff  & 64  &  5     &  6.20   &     1.07   &    0.93    &    0.55 \\
TopoDiff w/ G & 64  &  5   &     \textbf{5.44}    &   1.05   &   0.89   &     0.55  \\
\textbf{NITO (ours)} & 64   & 5   &        9.33     &   2.37  &    2.22   &     1.32  \\
\midrule
TopoDiff  & 64  &  10      &    2.91  &     0.71 &    0.64   &    0.33  \\
TopoDiff w/ G & 64  &  10  &    2.25   &    0.71  &   0.63  &     0.35  \\
\textbf{NITO (ours)} & 64   & 10    &   6.38  &    1.43   &   1.55    &    0.85  \\
\bottomrule
    \end{tabular}
    \label{tab:out-distro}
\end{table}
Something that should be looked at when it comes to these models is their performance generalizability to problems that are very different from the distribution of data used for training. To do this we test the performance of different models on an out-of-distribution test set for the 64x64 dataset. These results are presented in Table \ref{tab:out-distro} for TopoDiff and NITO. As it can be seen NITO's performance has significantly deteriorated on out-of-distribution data. This can be attributed to two matters. The first and most impactful is the nature of these models. TopoDiff generalizes better when it comes to these out-of-distribution tests given the fact that the model is generative in nature. This allows TopoDiff to handle out-of-distribution conditions better. To understand this we can look at generative models as a sort of retrieval approach, which allows the models to generate detailed and high-quality samples by finding the most similar topologies from the training data even when faced with very different inputs. On the other hand, NITO is a deterministic model that learns to map specific boundary conditions to near-optimal density fields. This makes it rather challenging for models like NITO when it comes to generating topologies for unseen and very different inputs since the mapping is a deterministic one. This highlights the importance of future work focusing on transforming our framework to be generative, which should be possible since many works have shown implicit neural approaches can be made generative~\cite{you2023generative}. The second potential reason contributing to TopoDiff's better performance on out-of-distribution samples is the use of physics-based fields for conditioning. This is because mapping stress fields to topologies makes it easier to handle very different boundary conditions since the stress fields may still be similar to samples in the dataset. However, this creates the kind of limitation that we discussed before, as such it is better to focus on making more robust generative schemes rather than using physical fields.
\begin{figure}[h]
    \centering
\includegraphics[width=\linewidth]{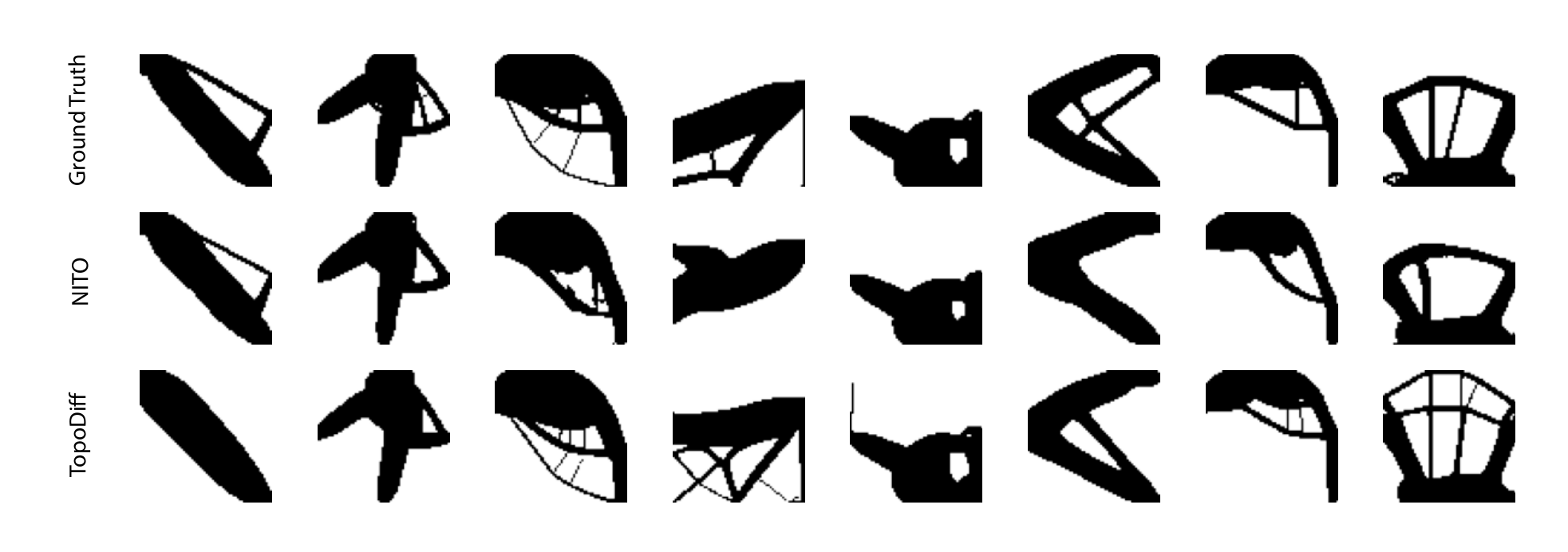}
\includegraphics[width=\linewidth]{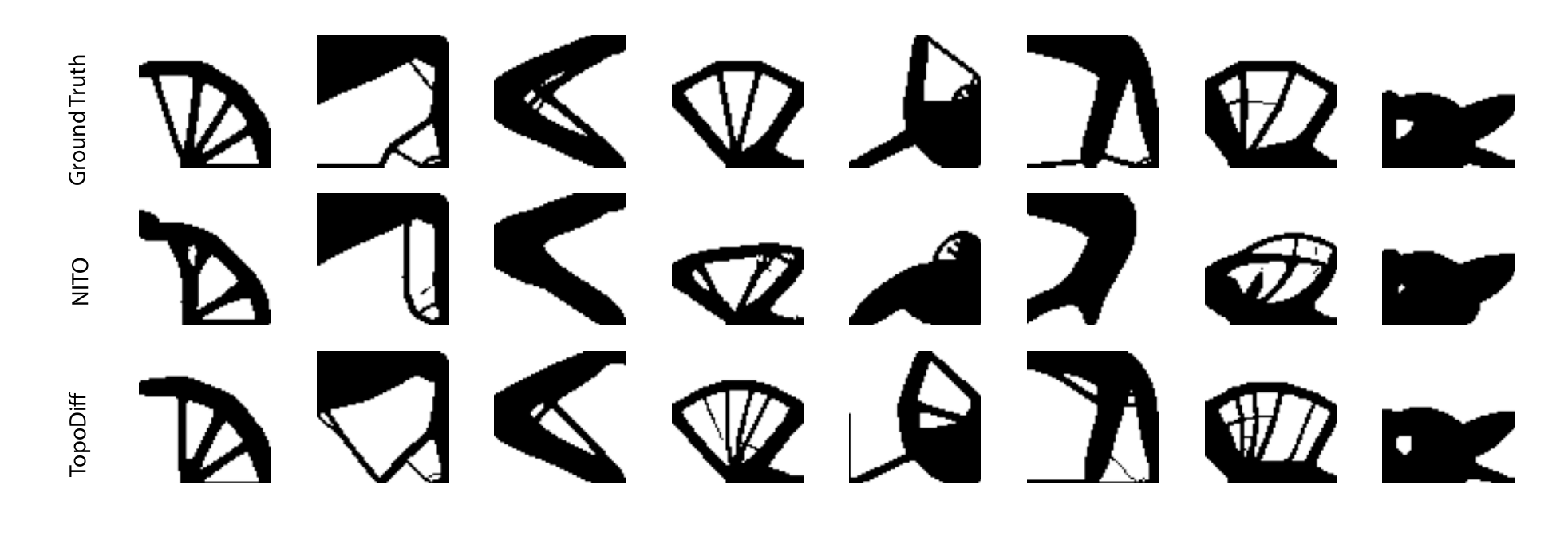}
    \caption{Visual comparison of samples generated for the out-of-distribution test. Each row is labeled. Ground truth samples are SIMP-optimized samples.}
    \label{fig:outdistro}
\end{figure}

\clearpage

\subsection{Further Examination Of Results \& Outperforming The Optimizer}
\begin{figure}[h]
    \centering
\includegraphics[width=0.7\linewidth]{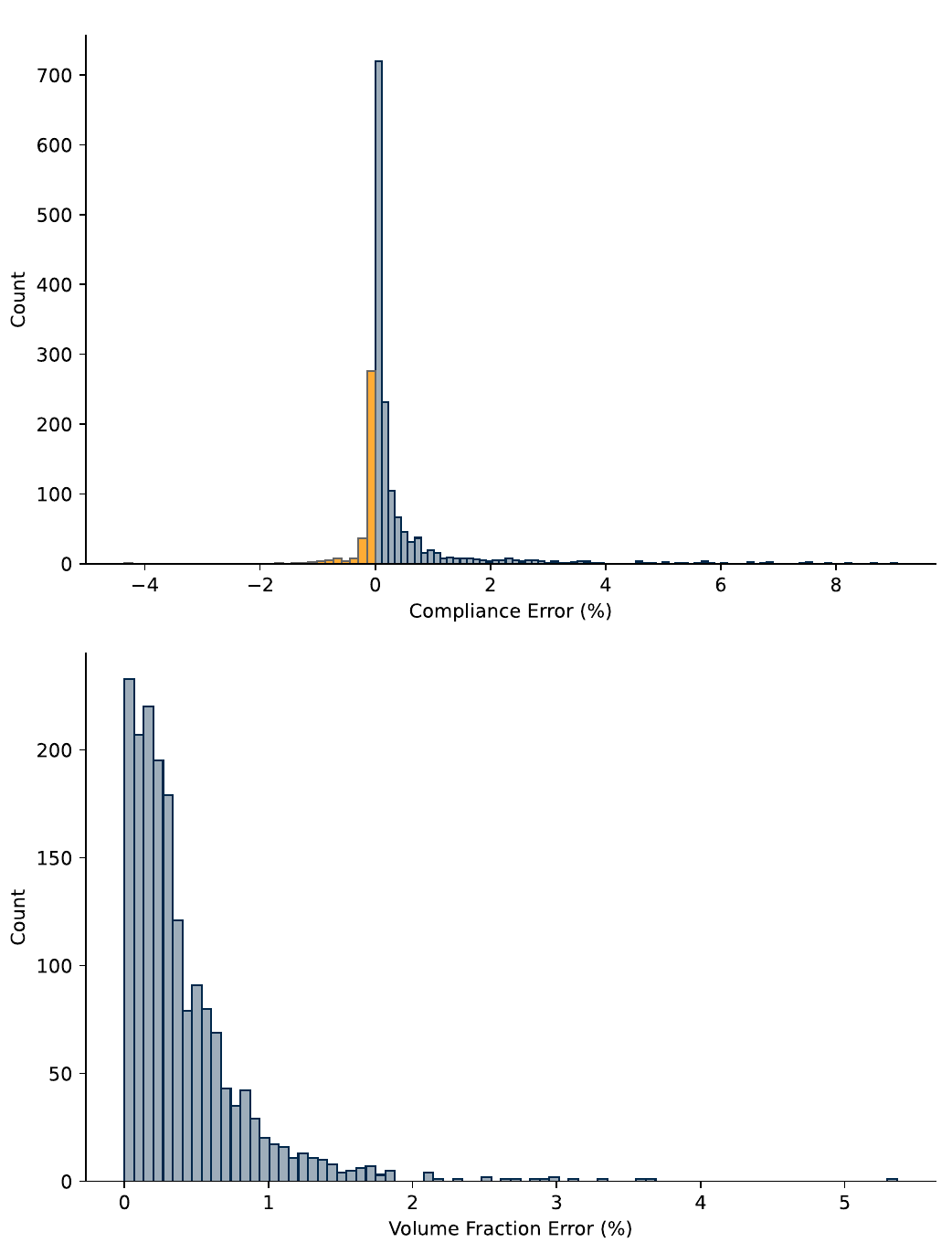}
    \caption{Distributions of volume fraction error (Bottom) and Compliance Error(Top). We see that NITO is capable of out-performing SIMP 19\% of the time as highlighted by yellow in the negative compliance errors~(meaning better than SIMP) on the top histogram. This data is for the 64x64 test set.}
    \label{fig:perfdistro}
\end{figure}

In Figure \ref{fig:perfdistro}, we show the distribution of compliance error and volume fraction error for NITO on the 64x64 dataset. We see that the majority of volume fraction error is below 1\% and a small number of high error samples skew the average. Similarly, we see that the majority of the compliance errors for NITO are below 1\% as well. Most notably, we observe that on 19.2\% of the samples in the test set NITO actually outperforms SIMP as indicated by the negative compliance errors visible on the histogram. This is a rather interesting outcome where NITO is capable of doing better than the optimizer that the training data came from. In Figure \ref{fig:bettertop}, we visualize a few instances of this phenomenon. It can be seen that in some problems NITO actually comes up with a different solution which as it happens out-performs SIMP, while in other instances the topologies are very similar and NITO has adjusted some of the finer details and redistributed the material differently to achieve better performance.

\begin{figure}[h]
    \centering
\includegraphics[width=\linewidth]{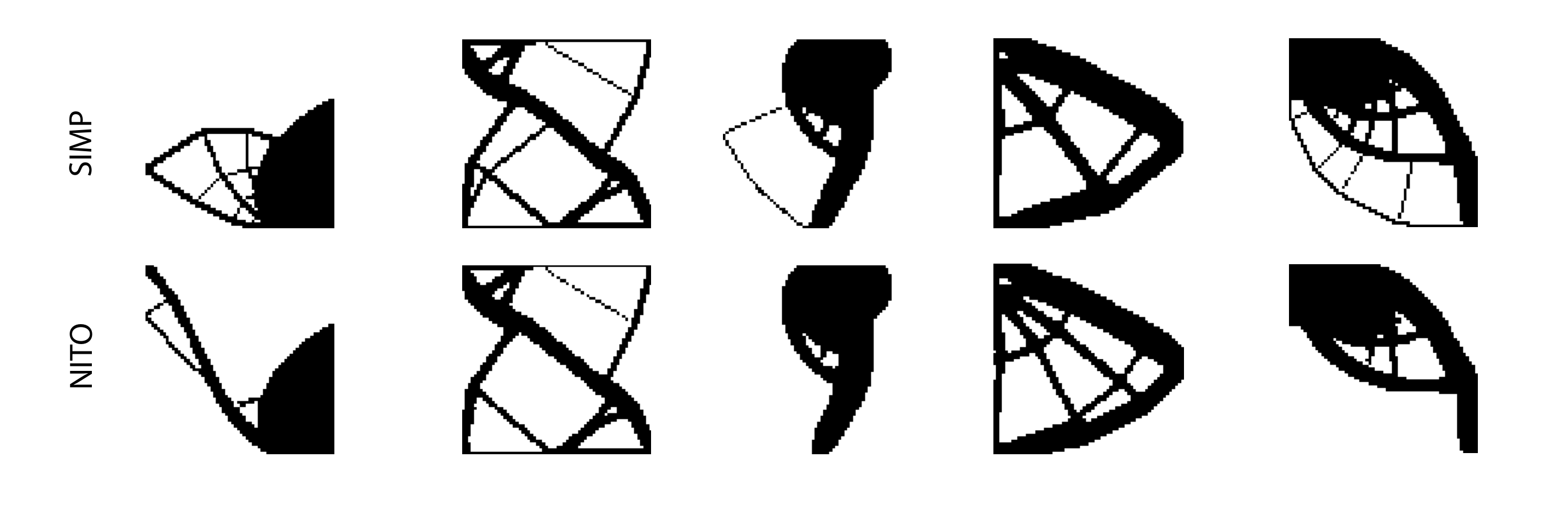}
    \caption{Example topologies where NITO outperforms SIMP. It can be seen that in some instances NITO finds very different topologies that outperform SIMP and in some instances, NITO has removed some details and redistributed the material in a way that has improved performance.}
    \label{fig:bettertop}
\end{figure}

\clearpage
\section{Visualizations}
\label{appx:visualizations}
Here we will provide a set of different visualizations for each test case.

\subsection{Direct Optimization Visualization}
In this section we provide some visualizations, demonstrating the effect of direct optimization on neural field-generated density fields. The main objective here is to showcase the importance and effectiveness of a few steps of direct optimization on the generated samples which completes the NITO framework. As can be seen in Figures \ref{fig:doptnito} and \ref{fig:doptnito256}, neural fields tend to have some smoothing and averaging in areas of high detail in the topology, which makes the baseline performance of neural fields worse. With only a small number of direct optimization iterations, however, we see that NITO can resolve the complex details effectively, showcasing why direct optimization is a crucial aspect of our framework.

In contrast, when we look at the effect of direct optimization on TopoDiff~(Figures \ref{fig:dopttopodiff} and \ref{fig:dopttopodiff256}) we see that in cases where the details predicted by TopoDiff are accurate, the optimizer does not provide much benefit, and when TopoDiff has not generated the correct details the generated topology is far from optimal which means that few steps of optimization does not provide significant benefits.

\begin{figure}[h!]
    \centering
\includegraphics[width=\linewidth]{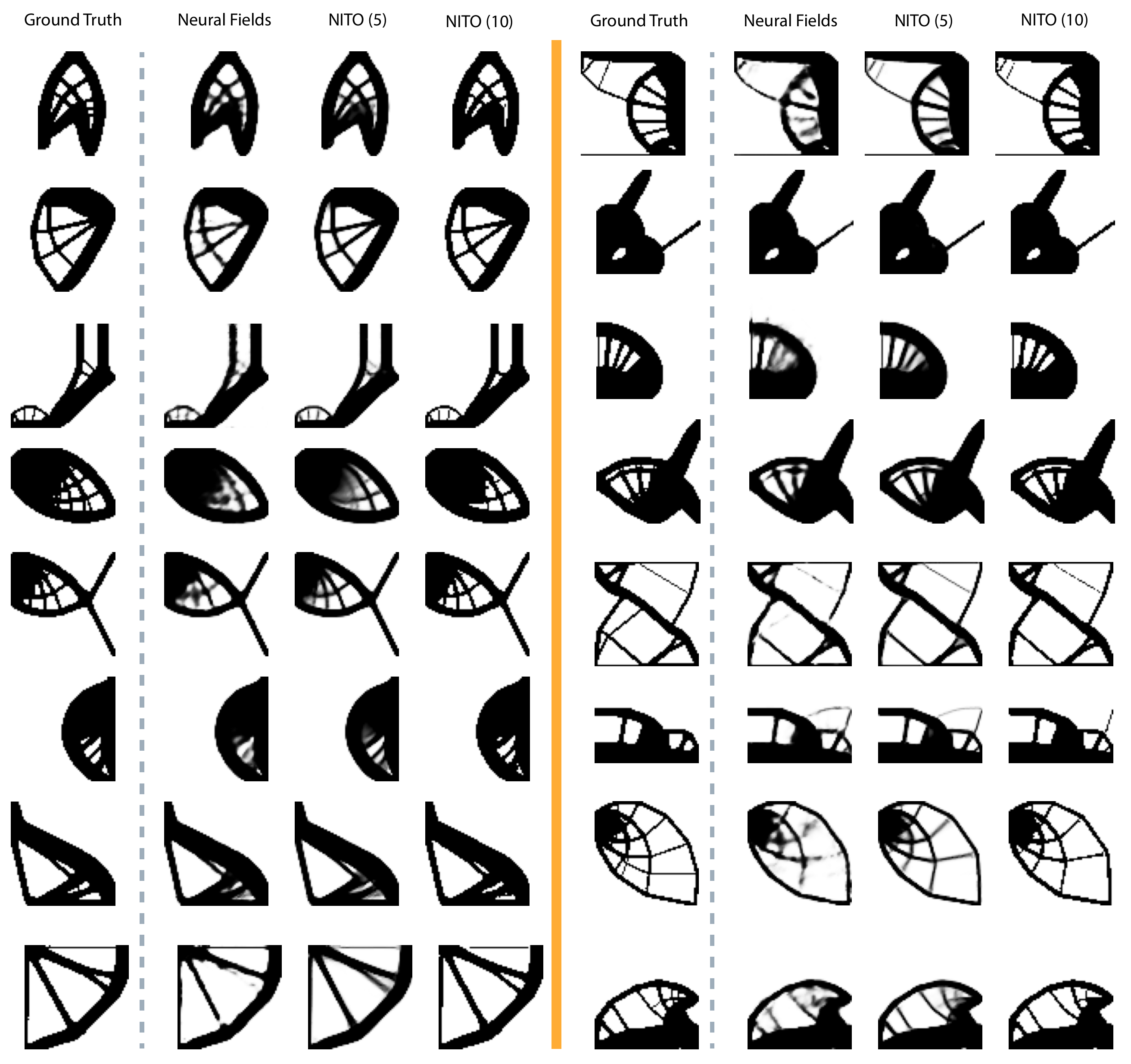}
    \caption{Visualization of the effect of direct optimization on samples generated by neural fields. The first column of each set of images shows the ground truth, the next column shows the raw predictions from neural fields, and the two columns that follow show the effects of 5 and 10 steps of direct optimization on the samples respectively. These samples are from the 64x64 test set.}
    \label{fig:doptnito}
\end{figure}

\begin{figure}[h!]
    \centering
\includegraphics[width=\linewidth]{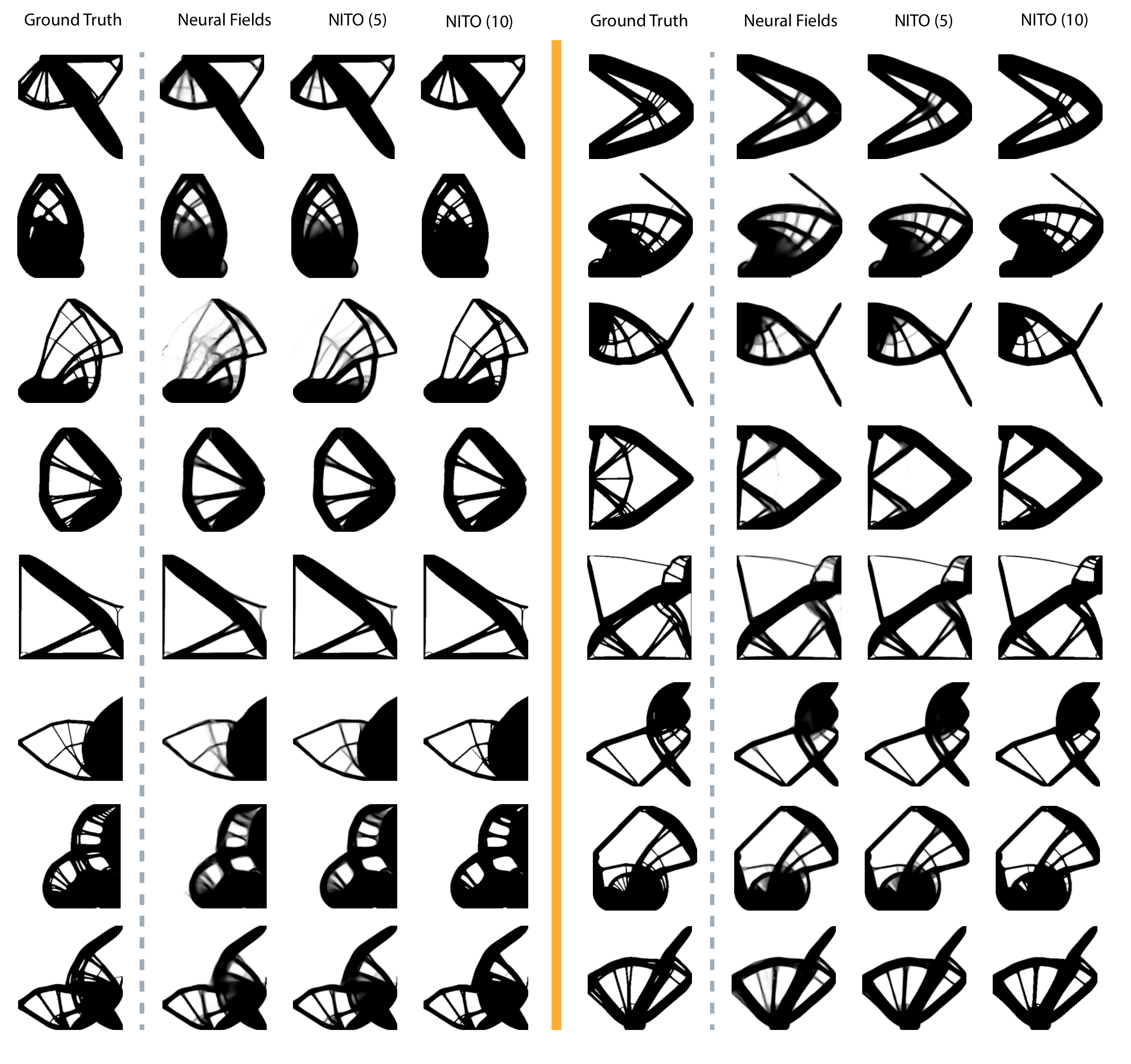}
    \caption{Visualization of the effect of direct optimization on samples generated by neural fields. The first column of each set of images shows the ground truth, the next column shows the raw predictions from neural fields, and the two columns that follow show the effects of 5 and 10 steps of direct optimization on the samples respectively. These samples are from the 256x256 test set.}
    \label{fig:doptnito256}
\end{figure}

\begin{figure}[h!]
    \centering
\includegraphics[width=\linewidth]{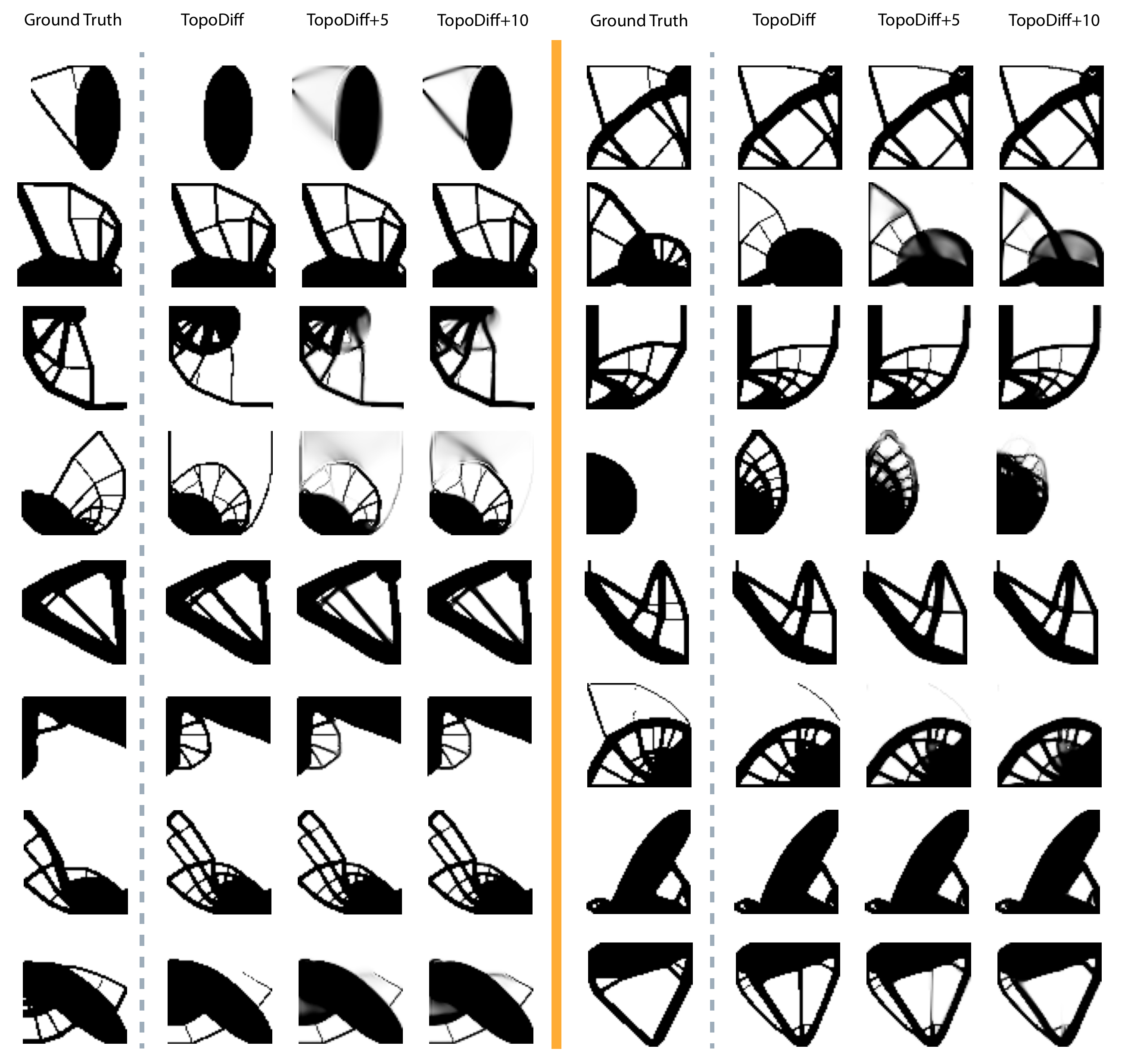}
    \caption{Visualization of the effect of direct optimization on samples generated by TopoDiff. The first column of each set of images shows the ground truth, the next column shows the samples generated by TopoDiff, and the two columns that follow show the effects of 5 and 10 steps of direct optimization on the samples respectively. These samples are from the 64x64 test set.}
    \label{fig:dopttopodiff}
\end{figure}

\begin{figure}[h!]
    \centering
\includegraphics[width=\linewidth]{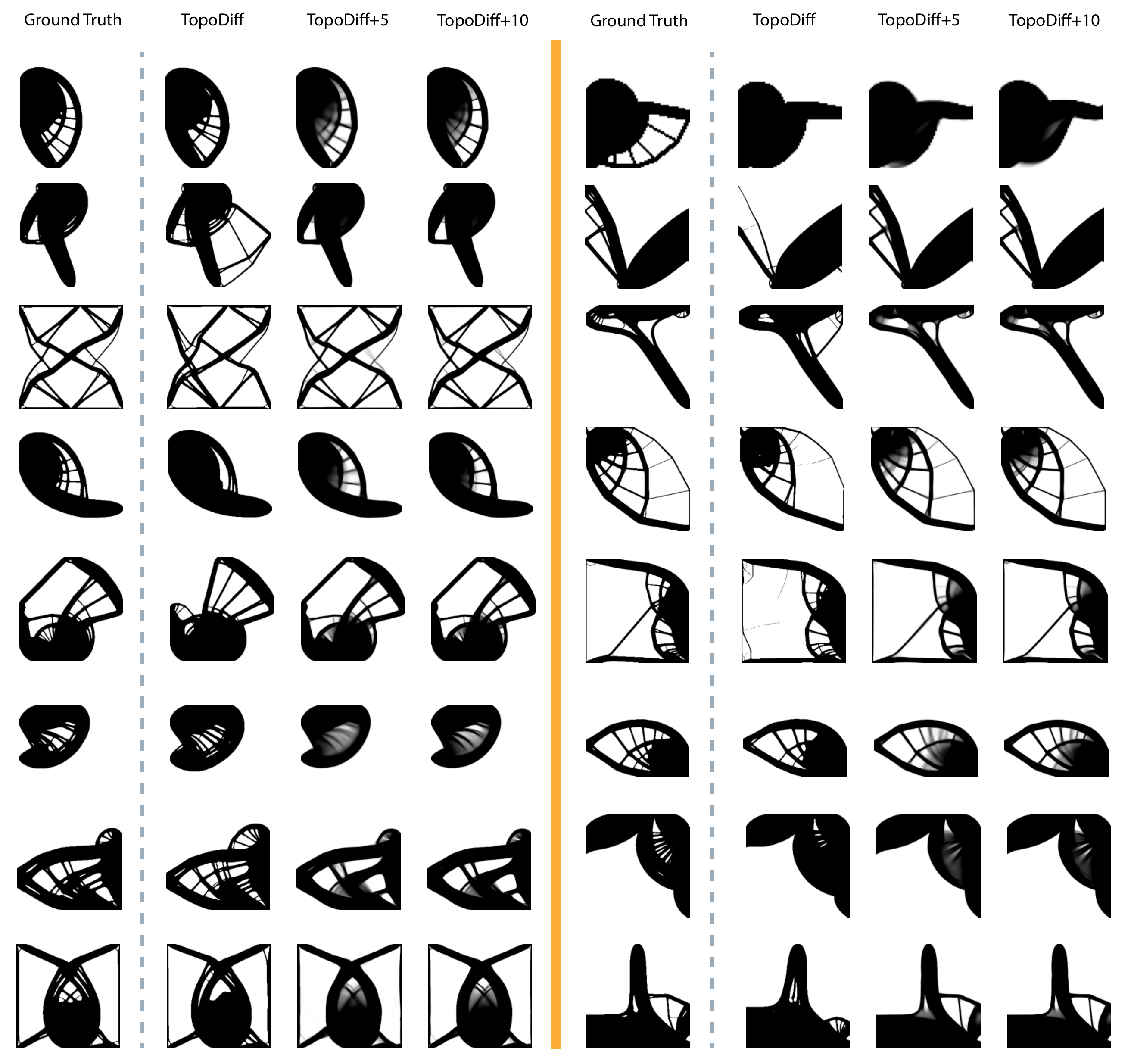}
    \caption{Visualization of the effect of direct optimization on samples generated by TopoDiff. The first column of each set of images shows the ground truth, the next column shows the samples generated by TopoDiff, and the two columns that follow show the effects of 5 and 10 steps of direct optimization on the samples respectively. These samples are from the 256x256 test set.}
    \label{fig:dopttopodiff256}
\end{figure}

\clearpage
\subsection{Generated Samples Visualizations}
In this section, we provide visualizations of samples predicted by different configurations of our model on different problems.

\begin{figure}[h!]
    \centering
\includegraphics[trim={0 11.5cm 0 11.5cm},clip,width=0.75\linewidth]{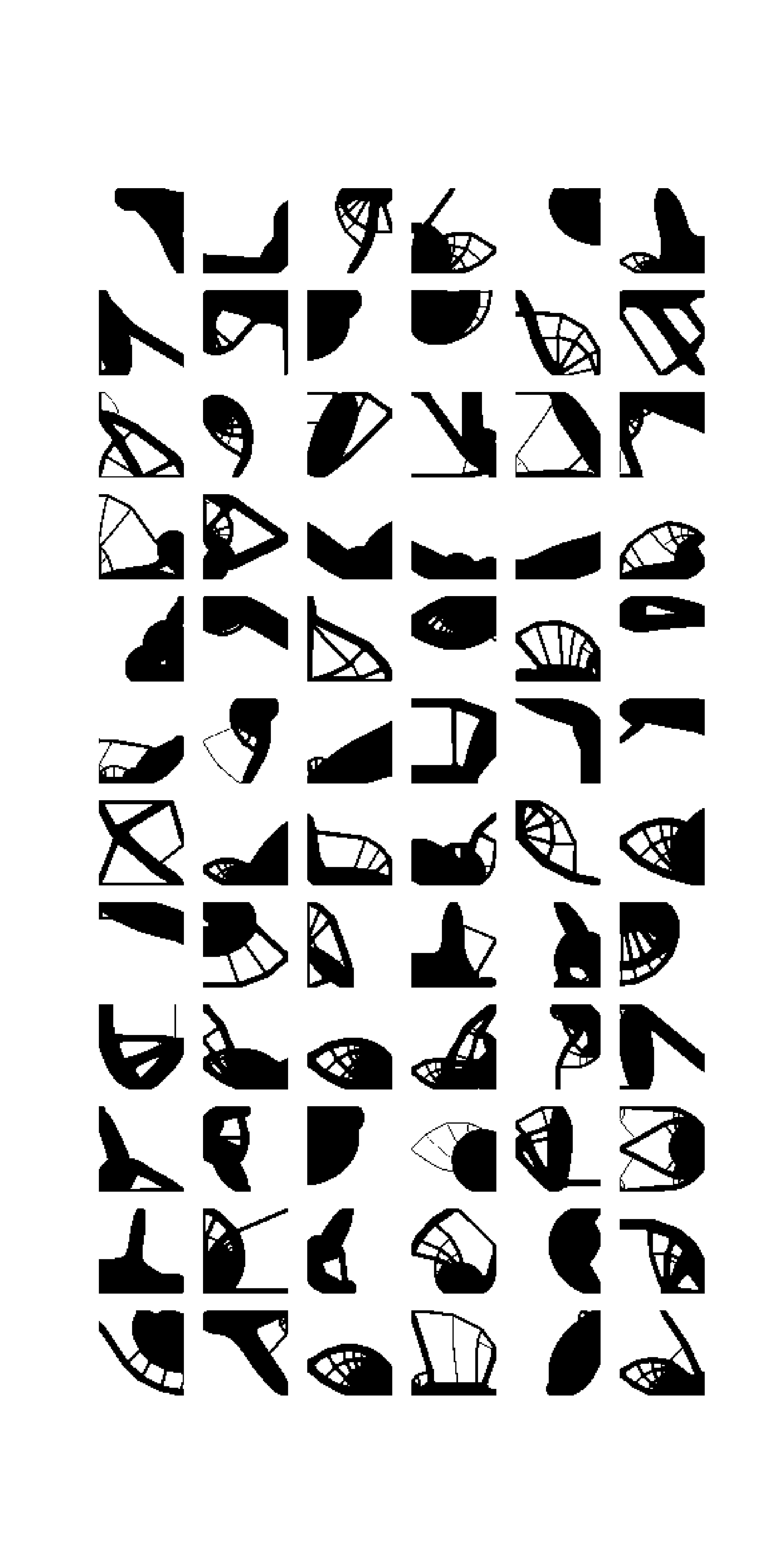}
    \caption{Ground truth images from the 64x64 SIMP datasets. Images that follow visualize NITO generated samples for the same problems.}
    \label{fig:64gt}
\end{figure}

\begin{figure}[h!]
    \centering
\includegraphics[trim={0 11cm 0 11cm},clip,width=0.75\linewidth]{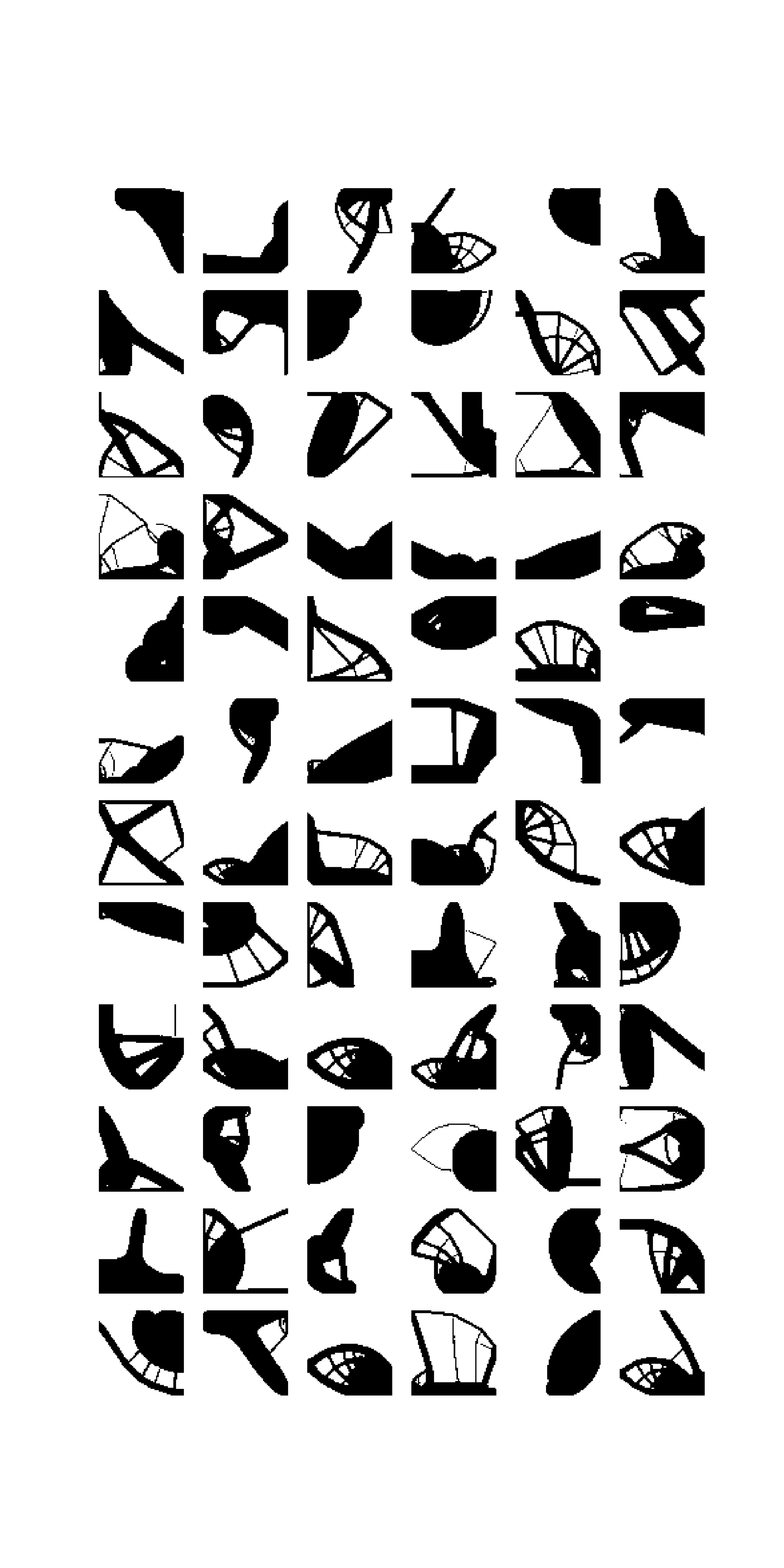}
    \caption{NITO generated topologies using a model trained on 64x64. Tested on the 64x64 data.}
    \label{fig:64gt}
\end{figure}

\begin{figure}[h!]
    \centering
\includegraphics[trim={0 11cm 0 11cm},clip,width=0.75\linewidth]{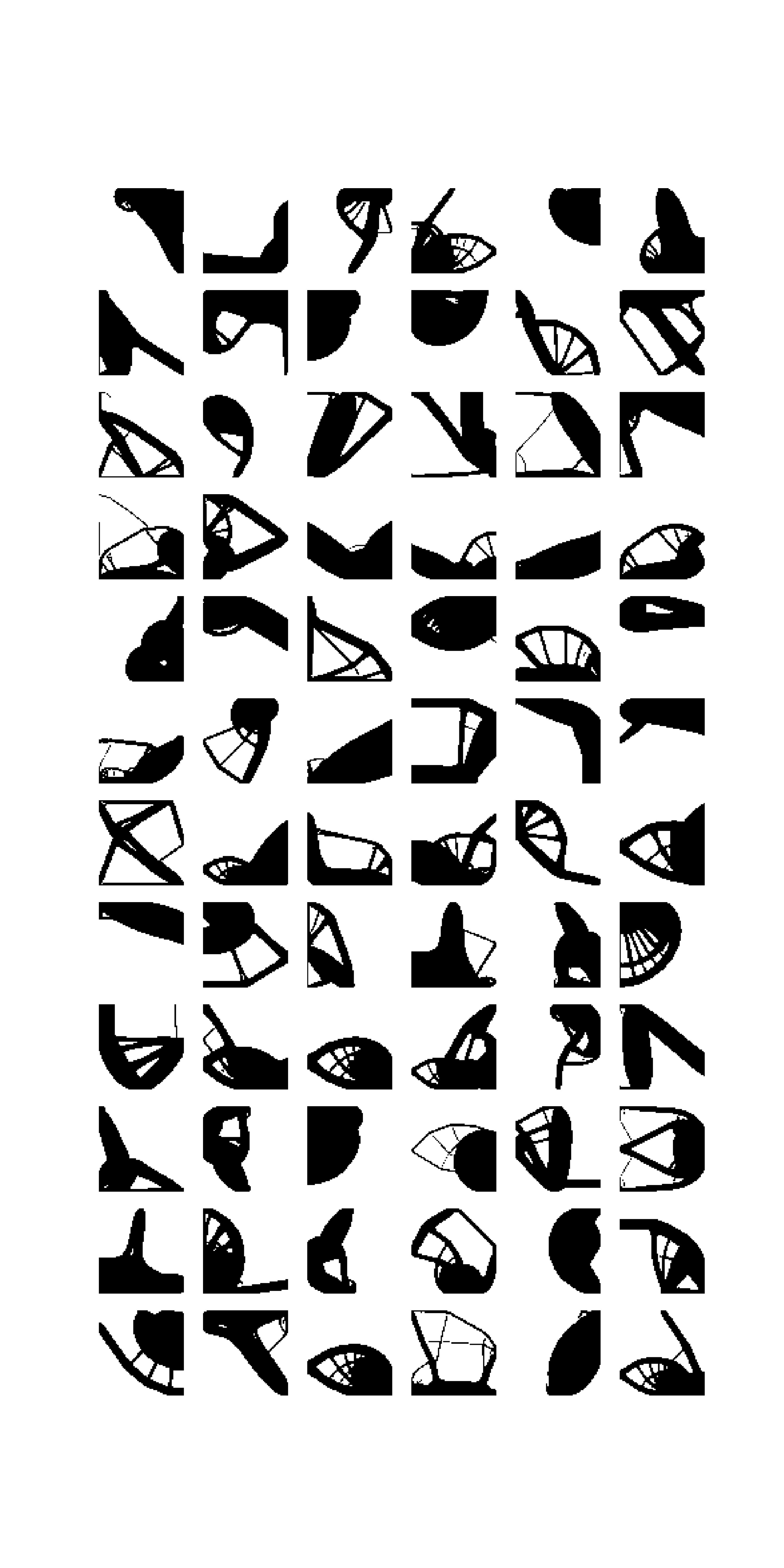}
    \caption{NITO generated topologies using a model trained on 256x256. Tested on the 64x64 data.}
    \label{fig:64gt}
\end{figure}

\begin{figure}[h!]
    \centering
\includegraphics[trim={0 11cm 0 11cm},clip,width=0.75\linewidth]{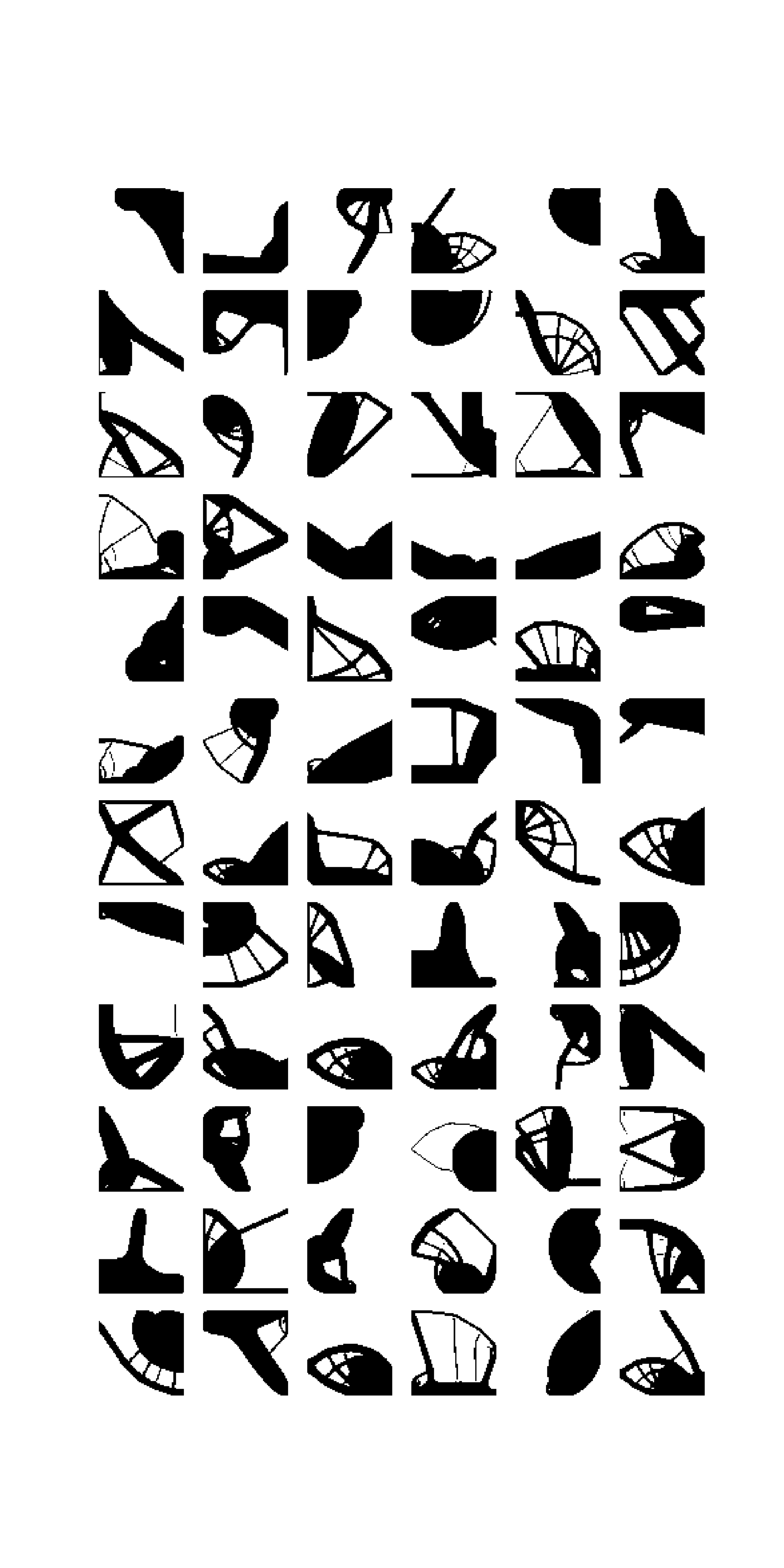}
    \caption{NITO generated topologies using a model trained on both 64x64 and 256x256. Tested on the 64x64 data.}
    \label{fig:64gt}
\end{figure}

\begin{figure}[h!]
    \centering
\includegraphics[trim={0 11.5cm 0 11.5cm},clip,width=0.75\linewidth]{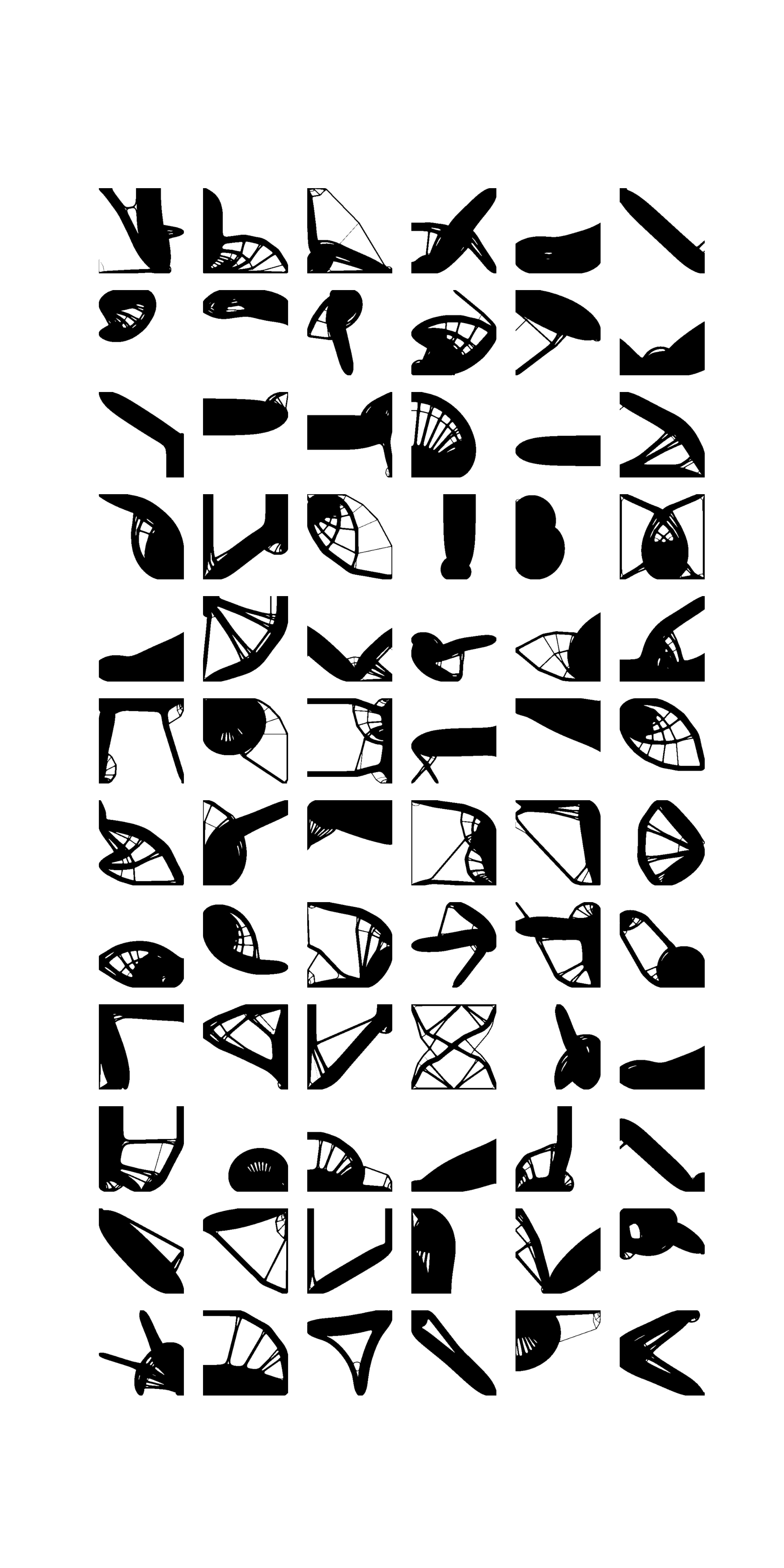}
    \caption{Ground truth images from the 256x256 SIMP datasets. Images that follow visualize NITO generated samples for the same problems.}
    \label{fig:64gt}
\end{figure}

\begin{figure}[h!]
    \centering
\includegraphics[trim={0 11cm 0 11cm},clip,width=0.75\linewidth]{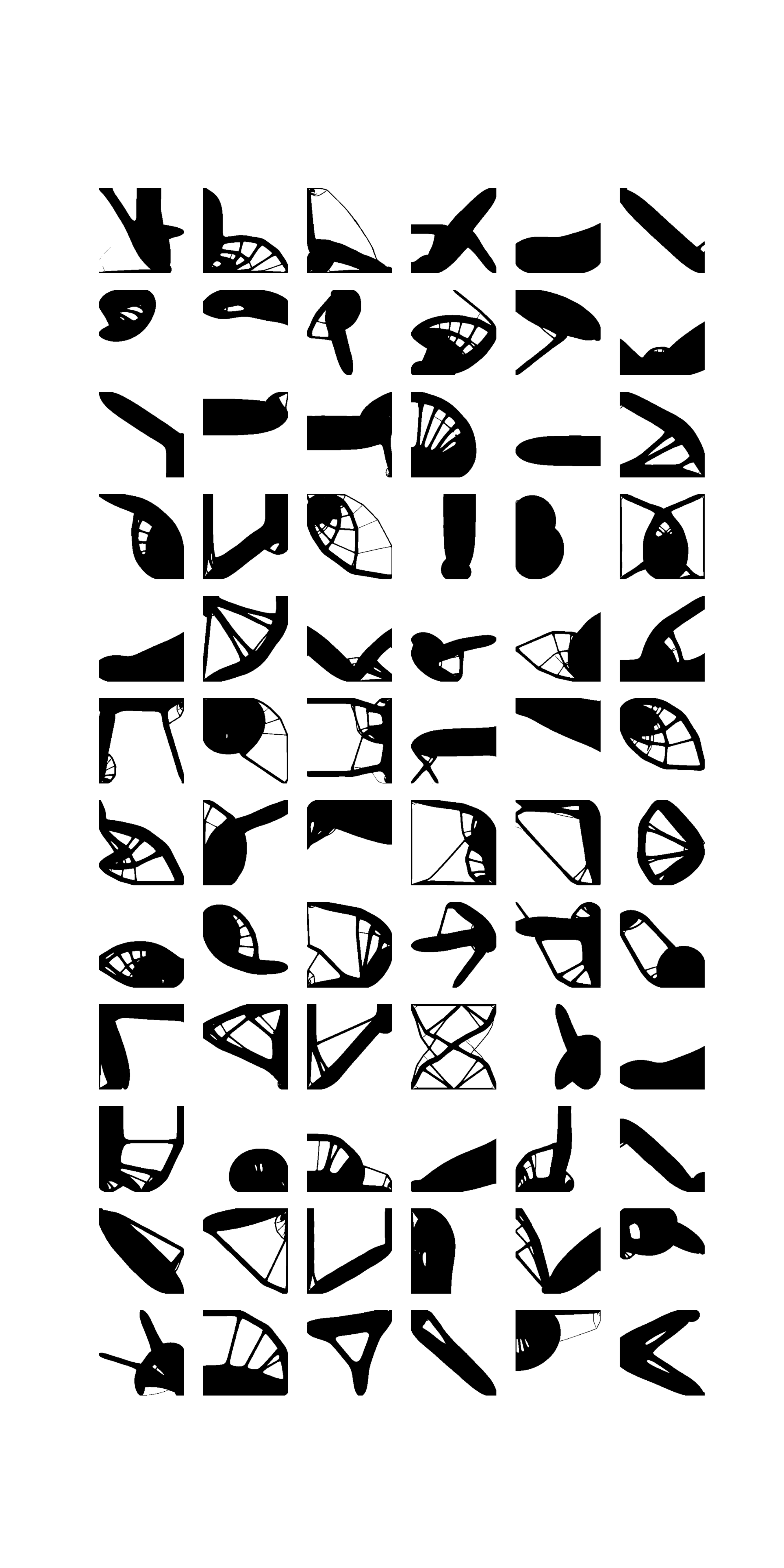}
    \caption{NITO generated topologies using a model trained on 256x256. Tested on the 256x256 data.}
    \label{fig:64gt}
\end{figure}

\begin{figure}[h!]
    \centering
\includegraphics[trim={0 11cm 0 11cm},clip,width=0.75\linewidth]{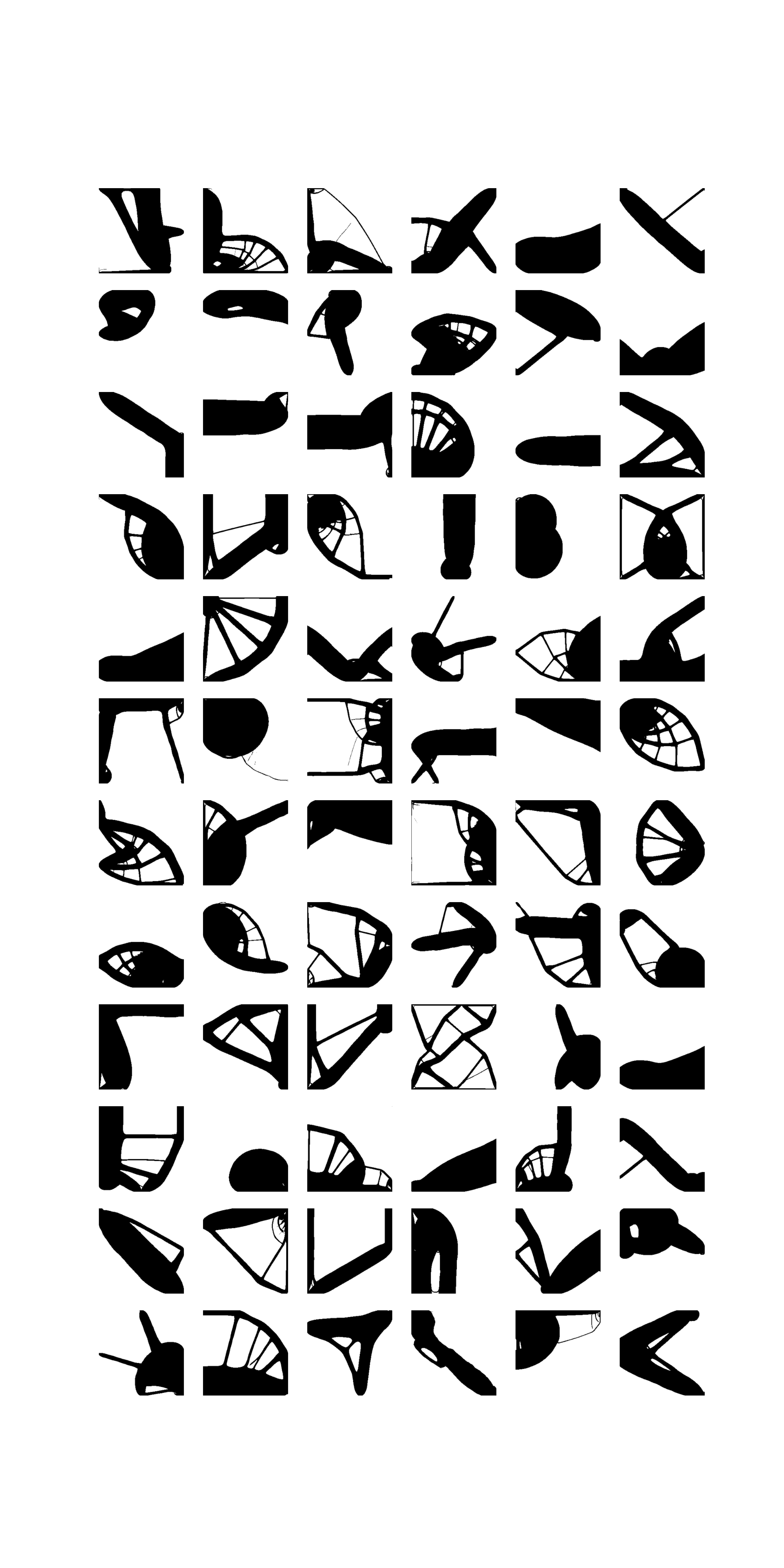}
    \caption{NITO generated topologies using a model trained on 64x64. Tested on the 256x256 data.}
    \label{fig:64gt}
\end{figure}

\begin{figure}[h!]
    \centering
\includegraphics[trim={0 11cm 0 11cm},clip,width=0.75\linewidth]{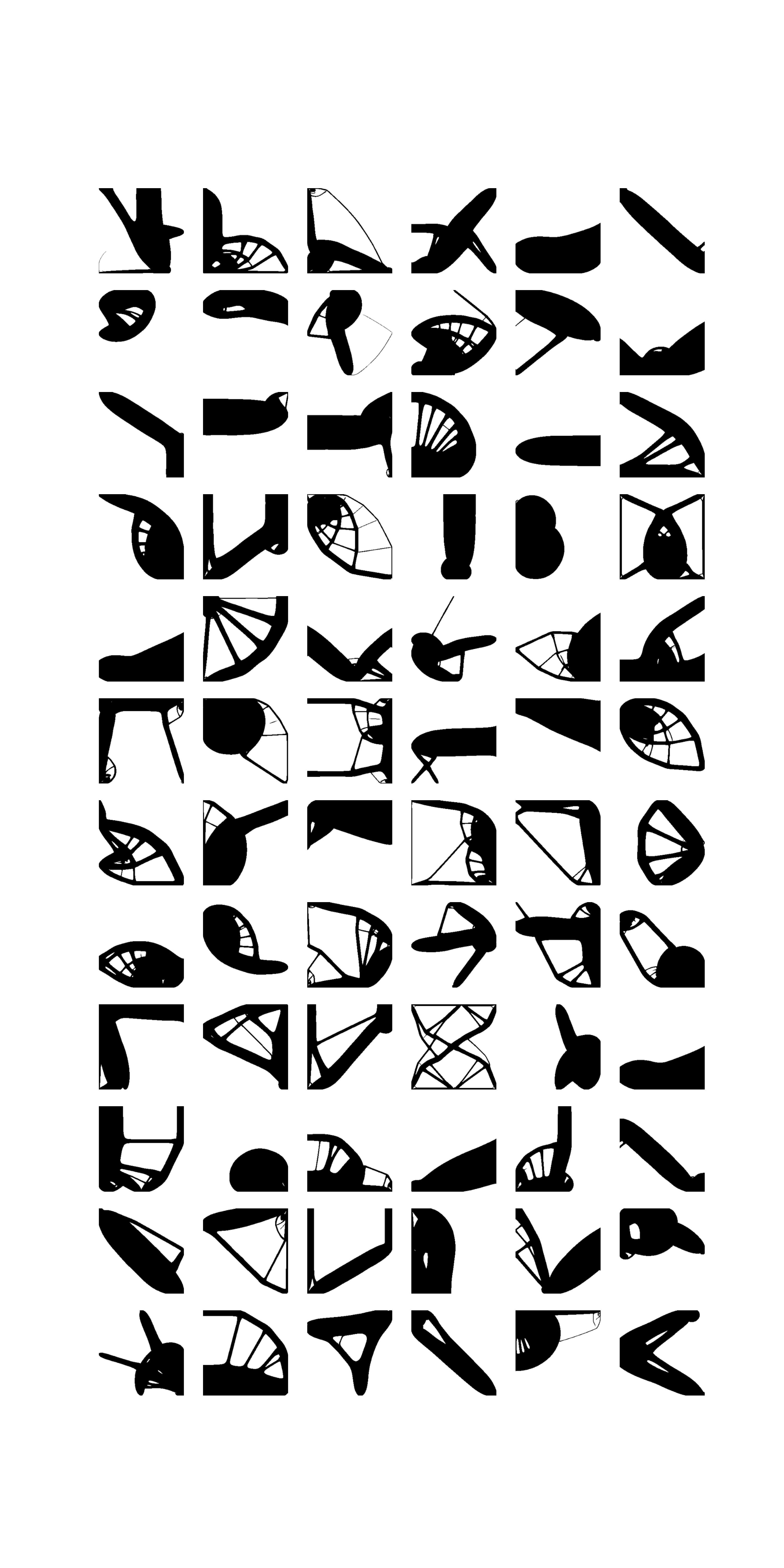}
    \caption{NITO generated topologies using a model trained on both 64x64 and 256x256. Tested on the 256x256 data.}
    \label{fig:64gt}
\end{figure}
\clearpage
\section{Implementation Details}

\label{appx:implementation-details}
Here we will briefly discuss some implementation details of the neural fields. First, the neural fields in our implementation are based on simple multi-layer preceptrons~(MLP), and amongst the different variations of implicit neural fields, we based our implementation on SIREN~\citep{siren} layers which apply sine activation functions to the output of each layer. Furthermore, it has been shown that implicit neural fields tend to ignore higher-frequency features and \citet{tancik2020fourier} have shown that transforming the spatial coordinates using Fourier feature mapping can be effective at mitigating this issue. As such we also apply Fourier feature mapping to the input coordinates of the model. 

Another important thing to note at this stage is the mechanism we use to incorporate the conditioning. Neural fields possess the capability to assimilate prior behaviors and exhibit generalization to novel fields through conditioning on a latent variable $\mathbf{C}$, which encapsulates the characteristics of a field. Specifically, this includes the boundary conditions and the volume ratio in our problem. \citet{perez2017film} propose Feature-wise Linear Modulation~(FiLM) which applies conditioning by modulating the outputs of the different layers of the model. To do this the conditioning mechanism includes two networks $\alpha(\mathbf{C})$ and $\beta(\mathbf{C})$, which predict a multiplicative and additive adjustment to the outputs of each layer. Our implementation is inspired by similar mechanisms built around normalization layers such as adaptive instance normalization~(AdaIN)~\citep{huang2017arbitrary}. Instead of modulating the outputs of each layer we apply layer normalization and modulate the scale and shift of the layer norm for each feature of the layer output. Putting this all together our neural field can be described as:
\begin{equation}
f_\theta(\mathbf{x}, \mathbf{C}) = f^{(L)} \circ f^{(L-1)} \circ \cdots \circ f^{(0)}(\mathbf{x}, \mathbf{C}))
\end{equation}
where $f^{(i)}$ for $i \in \{1,2,...,L-1\}$ indicates the function applied at each layer of the neural field except the first and last layer. Each layer in the model takes a hidden state $h^{(i)}$ as input and performs a linear transformation and layer normalization with modulation based on the condition vector:
\begin{equation}
\begin{split}
    f^{i}(h^{i},\mathbf{C}) = \sin(LN_{1,0}(W^{i} h^{i} + b^{i})\times  \alpha(\mathbf{C}) +\beta(\mathbf{C})),
\end{split}
\end{equation}
where $LN_{1,0}$ is layer normalization with scale 1 and shift zero and $\alpha$ and $\beta$ are fully-connected~(FC) layers that determine the feature-wise scale and shift based on the condition inputs. The first layer has the same function however the input coordinates are transformed by Fourier feature mapping. The final layer simply lacks the conditioning modulation layer normalization and $\sin$ activation function, instead, the activation function is $sigmoid$. As evident these neural fields are easily applicable to any domain shape as long as that information is contained in the condition latent representation $\mathbf{C}$. This is because the input to the neural field is simply coordinates, which allows this kind of density field prediction to be sampled arbitrarily in space, therefore, making this kind of modeling truly generalizable at least in theory to different domains, so long as the condition representation $\mathbf{C}$ is also generalizable. This is the topic is discussed in the main body of the paper. In our implementation, we use 8 layers of size 1024 for the neural fields and use 4 layers of size 256 for the point cloud models of which there are 3 one for force and one for constraints in each x and y direction. Further detail can be found in our code.

\subsection{Training Details}
The training of the conditional neural implicit model is performed in 3 stages. In the first stage, we perform training by sampling points of a 16x16 grid. That is to say that we sample points in space that each belong to one of the 16x16 divisions in space for the unit square domain. Furthermore, when sampling, if the material exists in a given sampling grid we randomly choose a point with material rather than picking points that are void. The first stage of training is carried out for 20 epochs. Followed by 20 epochs of sampling on a 32x32 grid and finally we sample on a 32x32 grid but rather than preferring material points we sample completely randomly. The final stage of training is 10 epochs. We use AdamW optimizer with a starting learning rate of $10^{-4}$ which is reduced on a cosine annealing schedule to be reduced at the end of each epoch to reach $5\times10^{-6}$ at the final epoch. The learning rate is stepped at the end of each epoch.

\clearpage
\section{Dataset \& Optimizer Details}
\label{appx:data}
Our experiments use a dataset of SIMP-optimized topologies in a unit square domain with a size of 64x64 and 256x256. For each topology in the dataset information about the loading condition, boundary condition, and volume fraction, are included. Furthermore, we include
the stress and strain energy fields. Our dataset is similar to the one proposed by ~\citet{maze2022topodiff}, however, we noted that the dataset used in prior works has been generated using an older version of SIMP, namely ToPy~\citep{hunter2017topy}. 

This solver, although a robust implementation of SIMP, does not implement the latest improvements to the SIMP algorithm and uses a slower solver, which causes two issues. Firstly, the topologies that are used in the dataset proposed by~\citet{maze2022topodiff} are lower-performing topologies in comparison to what the latest solvers produce, hence overselling the performance of these models. Secondly, the SIMP method itself that prior studies compared their inference time with did not use the fastest solvers, making those comparisons also inaccurate. To ensure that this is not the case in our studies we implement the SIMP optimizer from scratch in Python~(The code for which will be publicly available), which performs the optimization using the latest and fastest implementation of the SIMP algorithm as far as the authors are aware~\citep{WangZhaoZhou2021}. 

As such we re-create the 64x64 dataset proposed by ~\citet{maze2022topodiff} using our solvers and find that the resulting topologies using our method are significantly better performing in comparison to the prior datasets used by many other works of research~\citep{giannone2023aligning,giannone2023diffusing,maze2022topodiff,nie2021topologygan}. Given this, it is safe to assume that the performance of the models in prior studies that compare to the inferior dataset may have been overestimated. However, to allow for a fair comparison we retrain the best-performing model in the literature TopoDiff~\citep{maze2022topodiff} on our new dataset and rerun their experiments on this new dataset. However, we report the performance of other models as the original authors measured them.

The 64x64 dataset includes 30,000 training samples and 1,800 test samples which we use for testing our models. Additionally, we include a dataset of 1,000 out of distribution samples for out of distribution testing. The 256x256 dataset includes 60,000 training samples and 1,800 samples for testing. The figures that follow visualize some of the training samples for each dataset. In the following figures, we include figures showing samples from both datasets.

\begin{figure}[h]
    \centering
\includegraphics[width=\linewidth]{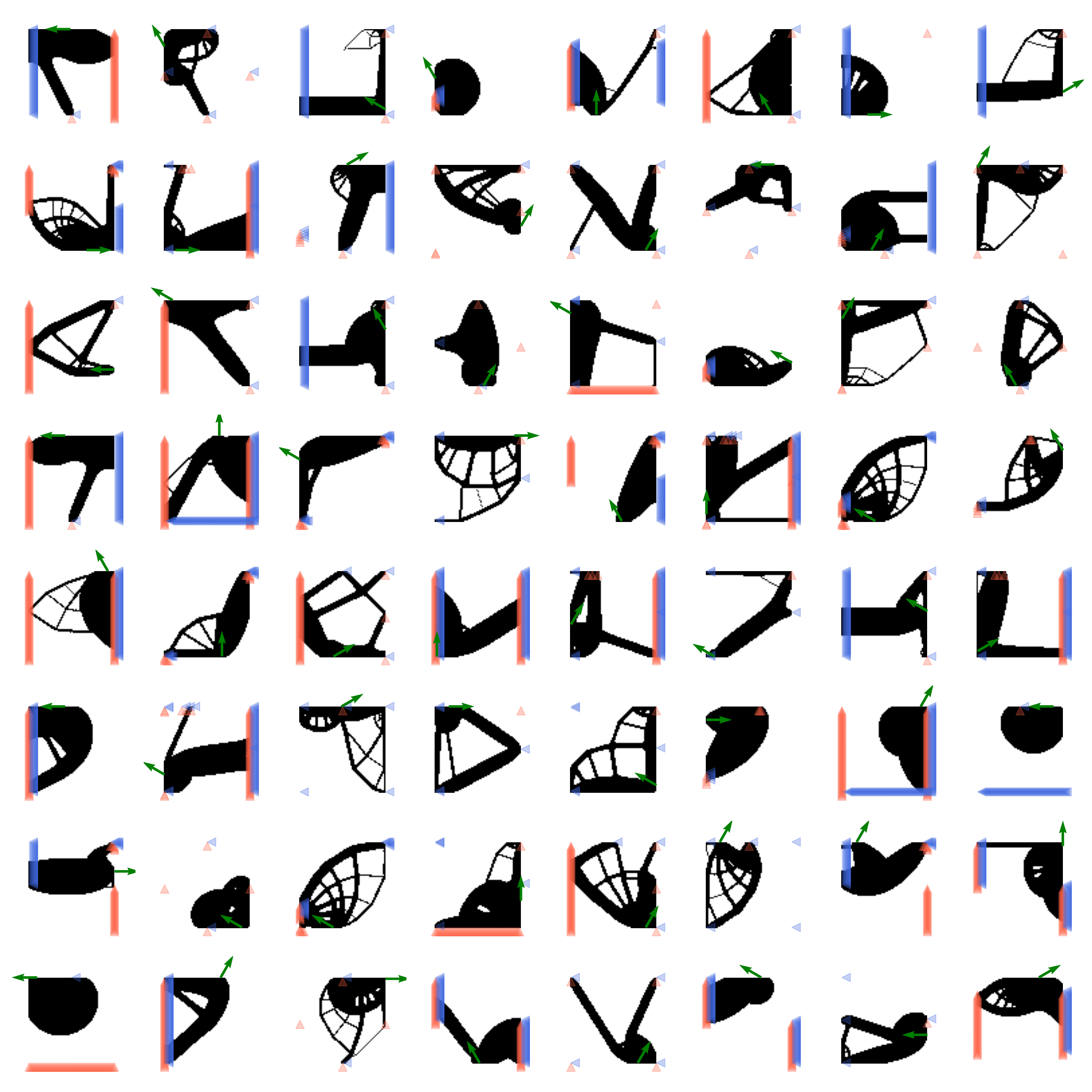}
    \caption{Random samples from the 64x64 dataset. Green arrows show the locations and directions of the loads applied in each problem. Blue triangles indicate points that are constrained in x and red triangles indicate points that are constrained in y.}
    \label{fig:bettertop}
\end{figure}

\begin{figure}[h]
    \centering
\includegraphics[width=\linewidth]{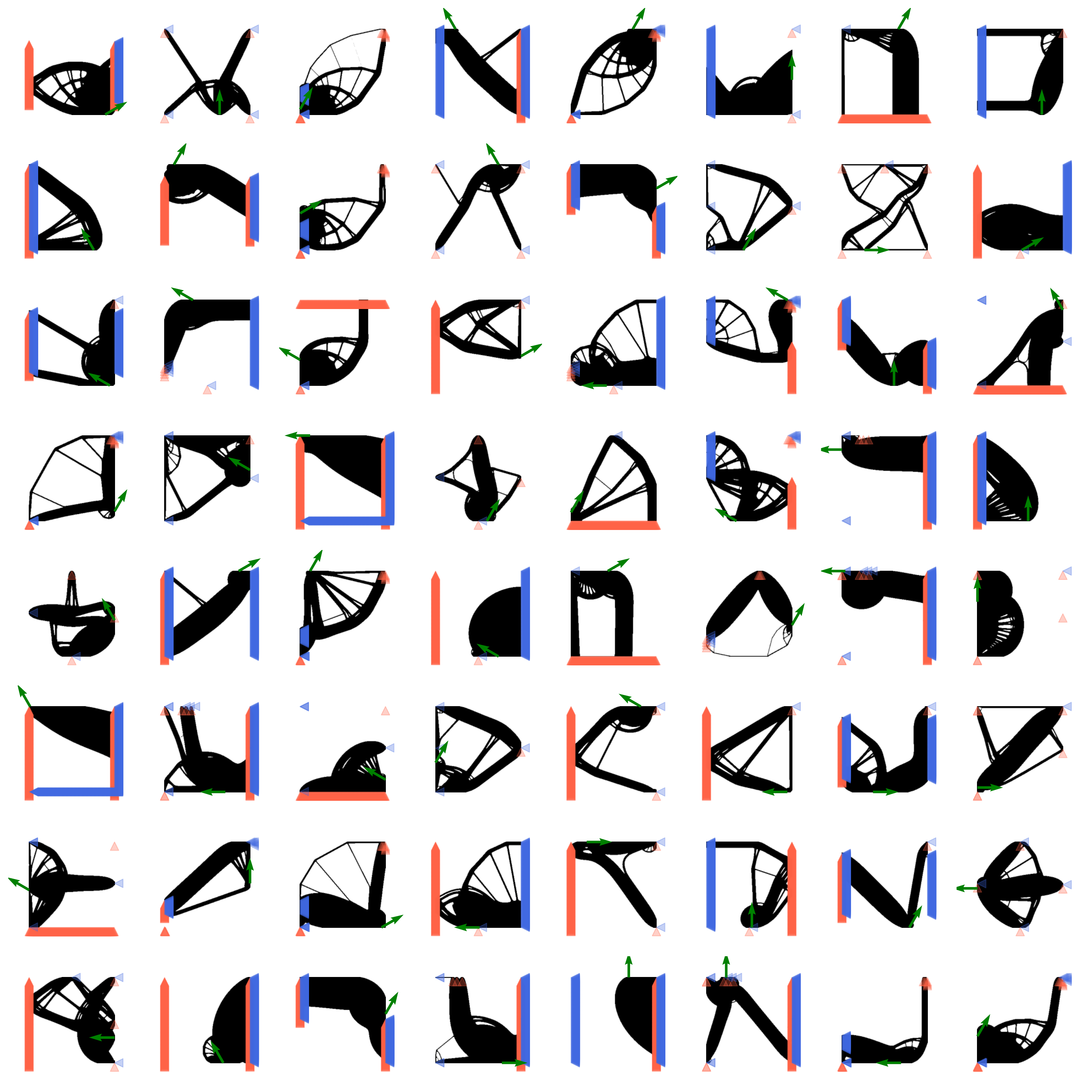}
    \caption{Random samples from the 256x256 dataset. Green arrows show the locations and directions of the loads applied in each problem. Blue triangles indicate points that are constrained in x and red triangles indicate points that are constrained in y.}
    \label{fig:bettertop}
\end{figure}

\end{document}